\documentclass[11pt]{article}

\usepackage[final]{acl}

\usepackage{times}
\usepackage{latexsym}
\usepackage{enumitem}

\usepackage[T1]{fontenc}

\usepackage[utf8]{inputenc}

\usepackage{microtype}

\usepackage{inconsolata}

\usepackage{graphicx}

\usepackage{float}
\usepackage[most]{tcolorbox}
\usepackage{booktabs}
\usepackage{amsmath,amssymb}
\newcounter{camealgorithm}
\newcounter{camealgline}
\usepackage[table]{xcolor}

\title{CAME: Company-Aware Evidence-Memory Experts for Interpretable Quarter-Ahead Revenue Forecasting}

\author{
  \textbf{Ya-Wen Wu},
  \textbf{Meng-Fen Chiang},
  \textbf{Kuang-Da Wang},
  \textbf{Wen-Chih Peng} \\
  National Yang Ming Chiao Tung University, Hsinchu, Taiwan \\
  \small{\texttt{yvonne0800111.cs13@nycu.edu.tw, meng.chiang@nycu.edu.tw}} \\
  \small{\texttt{gdwang.cs12@nycu.edu.tw, wcpeng@cs.nycu.edu.tw}} \\
}

\begin{document}
\maketitle
\begin{abstract}
Quarter-ahead revenue forecasting requires company-scale numerical accuracy, strict temporal validity, and company-specific interpretation of narrative disclosures. LLMs can distill textual evidence but can produce scale-misaligned forecasts, whereas history-based anchors are stable but miss forecast-time signals such as product transitions, supply constraints, and management guidance. We introduce \textbf{CAME} (\textbf{C}ompany-\textbf{A}ware Evidence-\textbf{M}emory \textbf{E}xperts), a residual-forecasting framework that refines a no-leakage statistical anchor when current semantic evidence and prior error patterns justify an adjustment. On a development-inclusive rolling backtest of 336 company-quarters from 12 large public technology and platform firms, CAME achieves the lowest aggregate point-estimate error among the reported methods, with statistically supported macro-sMAPE gains over the matched Statistical Anchor, and outperforms History + Guidance on all six aggregate metrics. CAME also links adjustments to source-linked evidence cards and guarded memory traces, supporting forecast inspection, provenance, and failure localization.
\end{abstract}

\section{Introduction}

Financial language models are increasingly evaluated on their ability to parse complex corporate disclosures, including regulatory filings and earnings transcripts. However, current benchmarks often prioritize intermediate tasks such as sentiment classification, financial question answering and numerical reasoning, or market-reaction proxies \citep{zhu-etal-2021-tat,chen-etal-2021-finqa,chen-etal-2022-convfinqa,reddy-etal-2024-docfinqa,shah2023zero,li-lu-2026-decoding}. While these tasks demonstrate linguistic competence, quarter-ahead revenue forecasting provides a distinct testbed for grounded numerical reasoning. Compared with directional or relative-outcome tasks and broader multi-fundamental forecasting settings \citep{koval-etal-2023-forecasting,koval-etal-2024-financial,divo2025forecasting}, our task targets an absolute quarter-ahead revenue level on each company's own scale. A robust forecast requires synchronizing forecast-time narrative signals (e.g., demand shifts, supply constraints) with a firm's specific segment structure and historical revenue dynamics. Furthermore, high-stakes finance demands accountability: stakeholders require clear provenance for numerical departures from historical expectations.

Direct numerical generation via LLMs allows conditioning on rich narrative context but does not explicitly enforce alignment with a company's revenue scale. This paradigm faces two primary failure modes:
(i) \textbf{Calibration Failure:} Free-form generation lacks anchoring to a company's specific revenue scale. Minor retrieval or prompt variations can yield mathematically implausible numerical fluctuations.
(ii) \textbf{Provenance Failure:} A generated revenue figure may reflect latent model priors, narrative inertia, or spurious text associations. Without an explicit evidence-to-adjustment trace, these influences are difficult to diagnose.

History-only statistical forecasting suffers from \textbf{Contextual Blindness}. While stable and effective at exploiting historical series \citep{hyndman2008automatic, LIM20211748,divo2025forecasting}, configurations restricted to lagged numerical history cannot leverage forecast-time signals (e.g., product ramps, inventory normalization, management guidance) that appear in text before manifesting in numerical lags. This leads to anchor instability: in the absence of explicit guidance or during business regime shifts, history-driven anchors may be directionally reasonable but numerically stale.

To address these limitations, we introduce \textbf{CAME} (\textbf{C}ompany-\textbf{A}ware Evidence-\textbf{M}emory \textbf{E}xperts), a residual forecasting framework that reformulates revenue prediction as an anchor-relative correction task. Rather than treating the LLM as an unconstrained forecaster, CAME asks a narrower, inspectable question: \textit{Given a scale-aligned statistical anchor, does current company-specific evidence justify a numerical refinement, and have similar historical patterns led this anchor to under- or over-predict?}
CAME is formulated as residual composition. A \textbf{Guidance-Gated Base Anchor} supplies the revenue reference: no-guidance rows can activate a guarded prior-error memory correction, while guidance-bearing rows preserve the statistical anchor behavior. A mixture of \textbf{Evidence-Memory Experts} then proposes residual refinements around this base, drawing on current \textit{evidence cards}, \textit{typed temporal memory}, and eligible numeric guidance. These proposals are composed with the base anchor through a shared anchor-relative aggregation step, preserving a traceable path from source-linked company evidence to quantitative adjustment while preventing narrative evidence from acting as an independent, uncalibrated forecaster.

CAME is company-aware in its representational design yet unified in its reasoning logic. All narrative evidence is interpreted relative to the target firm's specific revenue mechanisms (e.g., cloud-demand vs. advertising weakness). Furthermore, the framework decouples shared forecasting logic from declarative firm representations. By externalizing company profiles and segment schemas, CAME applies one shared forecasting procedure across the evaluated firms without requiring per-company model branching or fine-tuning.
We evaluate CAME on a development-inclusive rolling backtest of 336 instances for 12 large public technology and platform firms. Method and protocol selection used only AAPL, NVDA, and AVGO through FY2023 Q4; all settings were fixed before the temporal and company-held-out evaluations. FY2024 Q1--FY2025 Q4 forms a 96-row temporal held-out window, and nine additional firms form a company-held-out robustness panel. All firms use the same settings, without per-company tuning or method selection.

Our contributions are as follows:
\begin{itemize}[leftmargin=*]
\item We formalize quarter-ahead revenue forecasting as an anchor-relative correction problem, bridging the gap between narrative reasoning and scale-aligned numerical prediction.

\item We introduce CAME, a unified company-aware forecasting architecture that constructs base anchors from online revenue estimates using guarded prior-error memory, then applies expert-driven residual corrections.

\item We define a structured interface for interpretable forecasting. Source-linked evidence cards and forecast traces associate forecast adjustments with company segments, relation families, and temporally eligible memory cases.

\item We evaluate CAME on a development-inclusive rolling backtest of 336 instances for 12 large public technology and platform firms. CAME achieves aggregate gains over the matched statistical anchors and remains competitive with direct same-task LLM baselines, with the strongest within-panel evidence in no-guidance scenarios.
\end{itemize}

\begin{figure*}[!ht]
\centering
\includegraphics[width=0.92\textwidth]{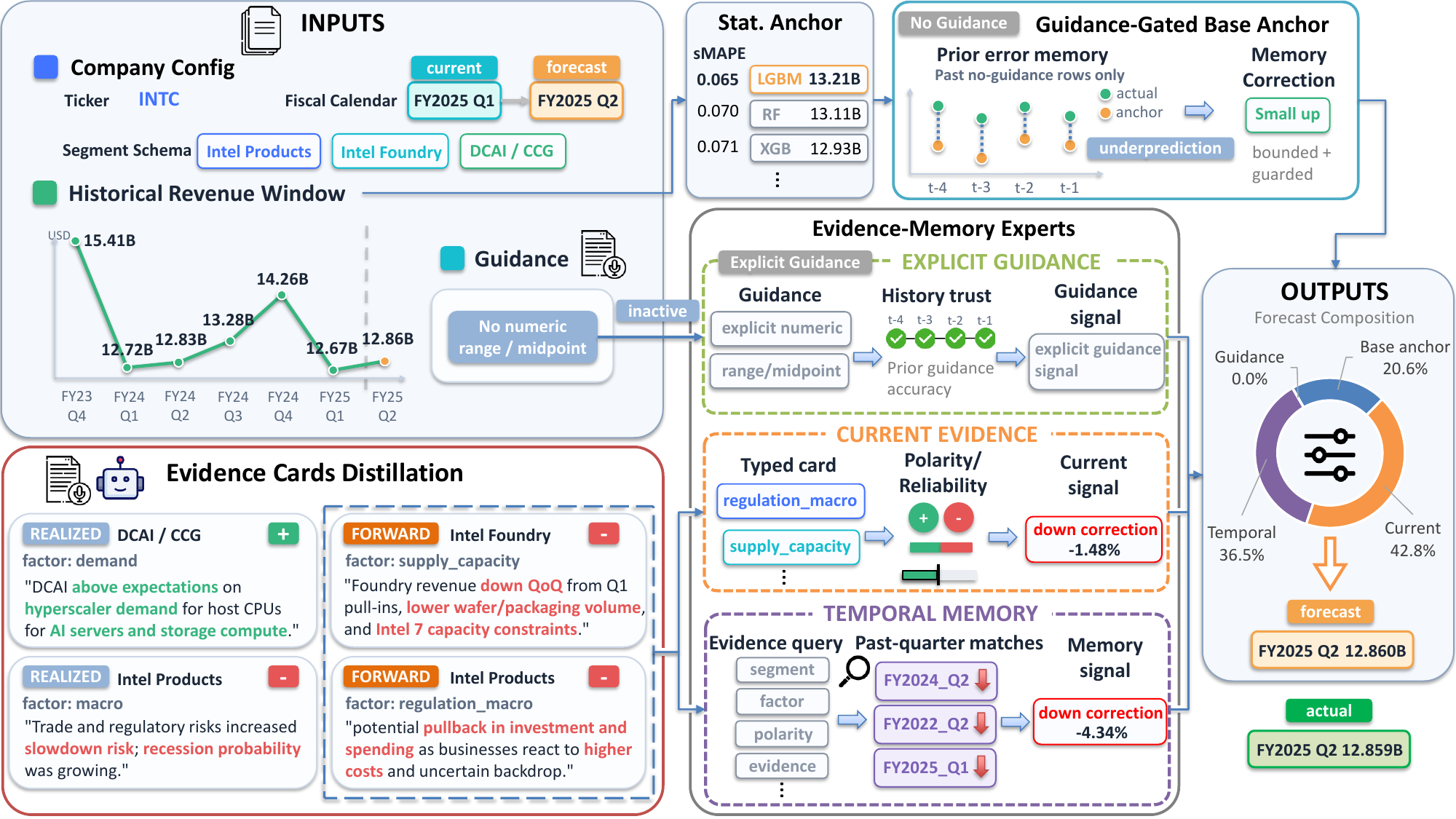}
\caption{\textbf{CAME Architecture.} CAME integrates company configuration, fiscal quarter, historical revenue, and temporally admissible narrative disclosures into typed, source-grounded \textit{Evidence Cards}. Starting from a \textit{Statistical Anchor}, CAME forms a \textit{Guidance-Gated Base Anchor}. Anchor-relative residual corrections are then composed from three \textit{Evidence-Memory Experts}: explicit guidance, current evidence, and temporal memory. CAME yields a scale-aligned revenue forecast with a source-linked trace that exposes adjustments and historical analogs for inspection.}
\label{fig:came-system}
\end{figure*}

\section{Related Work}

\paragraph{Financial Forecasting with Company Text.}
Financial NLP has studied predictive signals in filings, earnings calls, and financial reports for sentiment, risk, volatility, price movement, and earnings-related targets \citep{loughran2011liability,araci2020finbert,qin-yang-2019-say,sawhney-etal-2020-voltage,koval-etal-2023-forecasting}. Recent work moves closer to company-level forecasting by combining textual and tabular time series, comparing long financial reports, and evaluating statistical and machine-learning forecasters for fundamentals \citep{koval-etal-2024-financial,koval-etal-2024-learning,divo2025forecasting}. These studies show that company disclosures contain predictive information, but many targets are normalized surprises, risks, volatility, correlations, or market reactions. Quarter-ahead total revenue forecasting poses a different calibration problem: the prediction is an absolute revenue level on the company's own scale.

\paragraph{Numerical Reasoning and Retrieval.}
Financial QA and numerical reasoning benchmarks show that strong language models can struggle to combine narrative evidence, tables, and arithmetic consistently \citep{zhu-etal-2021-tat,chen-etal-2021-finqa,chen-etal-2022-convfinqa,reddy-etal-2024-docfinqa}. Sequential forecasting models and time-series LLMs provide tools for temporal patterns \citep{LIM20211748,nie2023a,jin2024timellm}, while retrieval-augmented and case-based reasoning methods support the grounded use of prior evidence \citep{10.5555/3495724.3496517,10.5555/196108.196115}. CARAG constructs a causal-temporal knowledge graph from financial statements and earnings calls, then uses specialist LLM agents to retrieve historical, guidance, and peer evidence for post-earnings price-shock prediction \citep{li-lu-2026-decoding}. These lines are related to CAME's use of temporally grounded evidence and memory, but our task requires estimating absolute quarter-ahead revenue on each company's own scale rather than directional market-reaction classification.

\paragraph{Traceability and Accountability.}
Interpretability in financial forecasting requires more than post-hoc feature attribution: users need to know which forecast-time evidence moved a numerical estimate and where conflicting evidence appears. CAME therefore treats source-linked evidence cards, memory cases, and correction traces as part of the forecasting interface.

\section{Problem Formulation}

\paragraph{Task Formalization.}
We define quarter-ahead revenue forecasting as a \textit{grounded numerical reasoning task}. Each instance is a tuple $(c_i, q_i, t_i)$, where $c_i$ denotes the target company, $q_i$ the target reporting quarter, and $t_i$ the forecast time (information cutoff).
The objective is to predict the realized total revenue $y_i\in \mathbb{R}^+$ by observing the information space $\mathcal{O}_i$ at time $t_i$. This space is partitioned into narrative disclosures $\mathcal{N}_i$ (e.g., transcripts, filings) and historical numerical series $\mathcal{H}_i$. The task is to learn a function $f$:
\begin{equation}
\hat{y}_i = f(c_i, q_i, \mathcal{O}_i), \quad \text{where } \mathcal{O}_i = \{ \mathcal{N}_i, \mathcal{H}_i \}.
\end{equation}

\paragraph{Forecast-Time Information Constraint.}
To preserve predictive integrity and prevent data leakage, we enforce a strict \textit{forecast-time information constraint}. For any instance $i$, the model's observation set $\mathcal{O}_i$ is restricted such that every element $j \in \mathcal{O}_i$ has a release time $r_j \le t_i$. Specifically, a retrieved historical memory case $m$ is eligible only if its realized revenue was observable by $t_i$.

\paragraph{Anchor-Relative Residual Forecasting.}
Directly generating raw revenue levels via LLMs can result in \textbf{scale misalignment}. We therefore reformulate the task as an \textbf{anchor-relative refinement}. Given a scale-aligned numerical prior $a_i$ (the Guidance-Gated Base Anchor), we define the target as the log-residual $\delta_i$:
\begin{equation}
\delta_i = \log(y_i) - \log(a_i).
\end{equation}
The model estimates $\hat{\delta}_i$, the multiplicative adjustment justified by forecast-time evidence. This formulation constrains the model to reason about \textit{relative change}, leveraging its linguistic capabilities, while delegating \textit{absolute scale} to the statistical anchor. Narrative evidence refines, rather than invents, the company's revenue scale, maintaining comparability with structured numerical baselines.

\section{Methodology}
We introduce \textbf{CAME}, a residual forecasting architecture that reconciles qualitative narrative signals with quantitative anchors.

\subsection{Framework Overview}
As depicted in Figure~\ref{fig:came-system}, CAME operates in four sequential phases:
(i) \textbf{Evidence-Card Distillation:} Narrative disclosures $\mathcal{N}_i$ are transformed into \textit{Evidence Cards}---typed semantic artifacts grounding raw text spans in business drivers, affected segments, and revenue-effect polarities.
(ii) \textbf{Guidance-Gated Base Anchor:} A no-leakage statistical family provides an initial revenue estimate $b_i$. For scenarios without target-quarter total-revenue guidance, CAME applies a guarded \textit{Anchor-Memory Correction} to $b_i$ to adjust for systematic historical biases, yielding the base anchor $a_i$.
(iii) \textbf{Evidence-Memory Experts:} A mixture of specialized experts proposes anchor-relative log-residuals $d_i^{k}$ by reconciling current evidence cards with temporal analogs (historical quarters with similar evidence-card configurations) and explicit numeric guidance.
(iv) \textbf{Guarded Composition:} Proposals are aggregated into a single residual $\hat{\delta}_i$ via reliability-weighted pooling. This phase applies a \textit{Guidance Guardrail} that linearly scales the revenue-space displacement from the base anchor according to the forecast-time guidance category.
Crucially, CAME's core reasoning logic is shared across the evaluated firms. Firm-specific nuances are decoupled from the architecture and externalized into declarative company profiles, fiscal-calendar metadata, segment and guidance schemas, and predefined relation sets.

\subsection{Company-Centered Evidence Cards}
To support source-linked inspection, CAME abstracts narrative context into structured semantic artifacts rather than processing raw documents during the forecasting phase. Formally, we define an extraction mapping $g_{\mathrm{ext}}(\mathcal{N}_i,c_i,q_i) \mapsto E_i$, where $E_i=\{v_{ij}\}$ is a set of source-linked evidence cards distilled from observable narrative disclosures $\mathcal{N}_i$ for target company $c_i$ and target quarter $q_i$.

These cards are not standalone forecasts; they are inspectable claims about company-specific revenue mechanisms, with typed fields for downstream experts. Each card $v$ $\in E_i$ is a tuple representing a source-grounded revenue mechanism:
\begin{equation}
{v = (x, h, \varsigma, \mathfrak{r}, p, s, \xi, \gamma, \rho),}
\end{equation}
where $x$ is the verbatim source span, $h$ is the affected temporal horizon, $\varsigma$ is the target-company segment, $\mathfrak{r}$ is the revenue-mechanism relation family (e.g., \textit{demand}, \textit{supply}), $p\in\{+, -, \pm, \mathrm{unk}\}$ is the revenue-effect polarity ($\pm$: mixed; $\mathrm{unk}$: indeterminate), $s$ is the signal strength, $\xi$ is the persistence hint, $\gamma$ is a heuristic extraction-confidence score rather than a calibrated probability, and $\rho$ is the functional role (e.g., \textit{realized} or \textit{forward-looking}).

CAME addresses divergent predictive implications of identical linguistic patterns across business models via declarative \textbf{Company Schemas} that map forecast-time language to normalized polarity and strength. Deterministic validators enforce predefined segment and relation sets, verify source-span availability, and enforce \textbf{temporal eligibility} under recorded release times by filtering information released after forecast time $t_i$. The extraction policy is designed to reject generic optimism and cost-side commentary unless it articulates a revenue driver; the human audit in Appendix~\ref{app:human-annotation-protocol} measures rather than assumes the semantic validity of this filtering. Validated cards then serve as inputs to the current-evidence expert, define retrieval keys for temporal memory, and gate the inclusion of explicit numeric guidance. (Appendix~\ref{app:evidence-interface} summarizes the evidence-card interface and extraction-prompt constraints.)
\subsection{Guidance-Gated Base Anchor}
The initial revenue prior is established by a statistical anchor $b_i$, drawn from a standardized family of forecasting models under a strict no-leakage, online protocol (see Appendix~\ref{app:anchor-expert-details} for model specifications). The Anchor-Memory Correction targets instances without target-quarter total-revenue guidance; these rows lack a management-provided numerical prior and show the largest matched-anchor errors among the evaluated guidance strata.

To mitigate these systematic errors, CAME constructs a \textbf{Guidance-Gated Base Anchor} $a_i$ by applying a guarded log-space memory correction $\Delta_i^{mem}$ to the raw anchor $b_i$. For observations without total-revenue guidance, the system evaluates the chronological sequence of prior no-guidance anchor residuals. If these historical errors satisfy a signed-strength admission gate, a bounded correction $m_i$ is synthesized to adjust the current prior; otherwise, the correction is null. Crucially, this branch is inactive for both explicit and non-explicit guidance-bearing instances, ensuring the anchor retains its original behavior whenever total-revenue guidance is present. By design, this mechanism functions as a correction layer for the base prior rather than as an additional narrative expert. We formalize this transition by defining a binary indicator $G_i^{mem}\in \{0,1\}$, which activates only when total-revenue guidance is absent and the admission gate passes:
\begin{equation}
\begin{aligned}
\Delta_i^{mem} &= G_i^{mem} \cdot m_i, \\
a_i &= b_i \exp(\Delta_i^{mem}).
\end{aligned}
\end{equation}

\subsection{Evidence-Memory Experts}
Given the base anchor $a_i$, CAME uses the expert-channel set $\mathcal{K}=\{\mathrm{cur},\mathrm{tmp},\mathrm{guid}\}$ to propose anchor-relative log-residuals $d_i^k$ with reliability scores $\sigma_i^k\in[0,1]$; inactive channels return zero reliability. These reliability scores are heuristic aggregation values, not calibrated probabilities or statistical confidence estimates. Shared prior rules transform them into pre-normalization masses $\tilde{\sigma}_i^k$. Let $\Omega_i=\sum_{k\in\mathcal{K}}\tilde{\sigma}_i^k$. The normalized weights and composed forecast are given as follows:
\begin{equation}
{
\begin{aligned}
w_i^k &=
\begin{cases}
\tilde{\sigma}_i^k/\Omega_i, & \Omega_i>0,\\
0, & \Omega_i=0,
\end{cases} \\
\hat{\delta}_i &= \sum_{k\in\mathcal{K}} w_i^k d_i^k, \\
\hat{y}_i^{\mathrm{pre}} &= a_i \exp(\hat{\delta}_i).
\end{aligned}
}
\label{eq:came-aggregation}
\end{equation}
When $\Omega_i=0$, all weights and the composed residual are zero, so CAME defaults to the base anchor. Each refinement module may therefore abstain when its underlying evidence is absent, redundant, or historically unreliable within the current guidance-availability context.

\noindent \textbf{Current-Evidence Expert.}
This expert estimates the log-residual justified by forecast-time narrative disclosures, distilled into \textit{evidence cards}. To prevent overfitting to signals already internalized by management guidance, we employ a \textit{partitioned learning target}: for no-guidance instances, the expert learns the full residual relative to $a_i$; for guidance-bearing instances, it learns only the residual unexplained by the matched path excluding this module. Both training and reliability estimation strictly adhere to the forecast-time information constraint, utilizing only prior realized observations.

\noindent \textbf{Temporal-Memory Expert.}
To overcome the limitations of lexically driven retrieval, this expert performs \textit{typed semantic retrieval} to identify historical analogs. It retrieves prior company quarters with similar evidence configurations (aligned across segments, revenue relations, and revenue-effect polarities), using their realized residuals as the basis for the current adjustment. This typed matching promotes functional relevance, mitigating noise from surface-level textual similarities (Appendix~\ref{app:anchor-expert-details} details retrieval weights).

\noindent \textbf{Explicit-Guidance Expert.}
This expert is narrowly scoped to reconcile the base anchor with the explicit numeric midpoints provided by management. It activates only when the statistical prior $b_i$ does not consume the current guidance signal, proposing a residual based on the log-ratio of the guidance midpoint and $a_i$. Its reliability is controlled by company-specific prior anchor-error history and current-row guidance-quality checks.

To support consistent inspectability, CAME applies shared aggregation logic across the evaluated firms. Conflicting evidence is exposed within the reliability-weighted trace rather than hidden behind company-specific forecast rules. In explicit numeric scenarios, the system automatically downweights current-evidence reliability to prioritize quantified management priors over qualitative interpretations, mitigating double-counting textual signals.

\subsection{Revenue-Space Guardrails}
CAME enforces numerical conservatism through a final \textbf{Guidance Guardrail} to the pre-guardrail forecast $\hat{y}_i^{pre}$ from \eqref{eq:came-aggregation}. This shared rule linearly scales the revenue-space displacement $\hat{y}_i^{pre}-a_i$ according to the forecast-time guidance category.

\paragraph{Revenue-Space Attenuation.}
We define a shared attenuation function that scales the revenue-space displacement from the base anchor $a_i$ to $\hat{y}_i^{pre}$ based on the forecast-time guidance category, partitioned into three regimes:
\begin{itemize}[leftmargin=*]
\item \textbf{Explicit Guidance or No Guidance:} Instances containing \textit{explicit numeric} guidance or falling under \textit{no-guidance} scenarios (where anchor memory may provide a guarded base-anchor correction) retain 100\% of the expert-weighted revenue-space displacement.
\item \textbf{High-Uncertainty Text:} For instances based on \textit{qualitative-only} or \textit{forward-looking commentary}, CAME retains 50\% of the revenue-space displacement to account for the ambiguity of non-quantified textual disclosures.
\item \textbf{Weak Signals:} Instances with \textit{weakly derived numeric} signals retain 0\% of the revenue-space displacement and therefore default to the base anchor $a_i$.
\end{itemize}
This revenue-space attenuation serves as an architectural constraint, preserving expert reliability scores, weights, and the composed log-residual while regularizing the output. Its coefficients are shared manual choices rather than per-company settings. Appendix~\ref{app:anchor-expert-details} reports one-factor sensitivity for the principal memory and retrieval settings; the guardrail coefficients are not varied.

\paragraph{Source-Linked Trace and Decision Provenance.}
CAME generates a deterministic source-linked trace for each of the 336 evaluated forecasts. Alongside the point estimate, the trace details linguistic and numerical provenance for inspection and failure localization. It documents:

\begin{tcolorbox}[
  colback=gray!8,
  colframe=black!85,
  boxrule=0.8pt,
  arc=2pt,
  left=6pt, right=6pt, top=6pt, bottom=6pt,
  title={\textbf{CAME Source-Linked Trace Components}},
  fonttitle=\small,
  coltitle=white,
  colbacktitle=black!85,
]
\small

\begin{enumerate}[label=(\alph*), nosep, leftmargin=10pt, itemsep=1pt, topsep=2pt]
\item The selected statistical prior and the activation status of the \textbf{Anchor-Memory} branch.
\item The normalized \textbf{Expert Weights} ($w_i^{k}$) and their respective heuristic reliability scores.
\item The specific \textbf{Evidence Cards} (supporting and conflicting) distilled from narrative disclosures.
\item The \textbf{Temporal Analogs} (historical company-quarters) retrieved for residual refinement.
\end{enumerate}
\end{tcolorbox}
The trace exposes the recorded expert composition and final forecast alongside source-linked evidence and historical analogs, supporting inspection and failure localization without establishing causal correctness or validated trace utility.

\section{Experiments}

\subsection{Experimental Setup and Design}

\noindent \textbf{Datasets.} We evaluate on a 12-company panel (AAPL, NVDA, AVGO, AMZN, ASML, GOOGL, INTC, META, MSFT, MU, TSLA, ORCL), comprising 336 company-quarter observations. AAPL, NVDA, and AVGO serve as development companies, with data through FY2023 Q4 used for method and protocol selection. FY2024 Q1--FY2025 Q4 constitutes the 96-row temporal held-out window. The remaining nine companies form the company-held-out robustness panel, untouched during method or protocol selection. All model-selection and protocol choices were fixed before the temporal and company-held-out evaluations; CAME is applied uniformly, without per-company tuning or method selection. (Appendix~\ref{app:evaluation-surfaces} details evaluation surfaces.)

\noindent \textbf{Comparative Baselines.} The primary statistical comparator is the Statistical Anchor (\textbf{SA}), matched row-by-row to CAME's pre-correction reference under an identical online no-leakage anchor-selection policy, evaluated before any guidance-gated memory or residual evidence corrections. We also compare against direct same-task LLM baselines and exact-coverage literature-grounded text comparators: wide-context revenue prompting, CARAG-style transcript retrieval, zero-shot transcript prompting, and filing-only / Management's Discussion and Analysis (MD\&A)-only configurations \citep{koval-etal-2023-forecasting,ni2024harnessing,li-lu-2026-decoding,shah2023zero}. These are all adapted for quarter-ahead total-revenue prediction on aligned company-quarter targets under consistent forecast-time information boundaries. The main table includes these exact-coverage text comparators; document-only variants serve as weaker controls. The SA is the primary same-task control, selected from a fixed family of recent-history, time-series, and guidance-aware anchors, with trainable candidates that use only prior rows and are selected based on prior realized errors. Two exact-coverage history-only time-series baselines are TimeGPT-1 and Chronos-Bolt Base \citep{garza2024timegpt,ansari2024chronos}; both provide 336/336 one-step coverage without guidance or text. The direct same-task LLM baselines are History-only and History + Guidance. History-only uses prior realized quarterly revenue, whereas History + Guidance additionally receives prior guidance and available numeric target-quarter guidance. When numeric target-quarter guidance is unavailable, History + Guidance reuses the corresponding History-only forecast. Appendix~\ref{app:prompt-comparator-details} provides the exact prompts, company-label treatment, fallback protocol, and masking limitations.

\begin{table*}[!ht]
\centering
\footnotesize
\setlength{\tabcolsep}{3pt}
\resizebox{\textwidth}{!}{%
\begin{tabular}{lrrrrrrr}
\toprule
\textbf{Method} & \textbf{N} & \textbf{sMAPE $\downarrow$} & \textbf{MAPE $\downarrow$} & \textbf{MAE $\downarrow$} & \textbf{RMSE $\downarrow$} & \textbf{$R^2$ $\uparrow$} & \textbf{DA $\uparrow$} \\
\midrule
\rowcolor{gray!8} \multicolumn{8}{l}{\emph{Adapted Text and Retrieval Baselines}} \\
Wide-context revenue & 336 & 0.071 $\pm$ 0.031 & 0.072 $\pm$ 0.034 & 3.04B $\pm$ 3.87B & 4.13B $\pm$ 5.41B & 0.876 $\pm$ 0.136 & 79.3\% $\pm$ 14.2\% \\
CARAG-style retrieval & 336 & 0.083 $\pm$ 0.036 & 0.086 $\pm$ 0.044 & 2.98B $\pm$ 3.52B & 4.03B $\pm$ 4.89B & 0.837 $\pm$ 0.176 & 81.2\% $\pm$ 9.3\% \\
\midrule
\rowcolor{gray!8}\multicolumn{8}{l}{\emph{Document-Only LLM Baselines}} \\
Transcript-only & 336 & 0.107 $\pm$ 0.053 & 0.119 $\pm$ 0.072 & 3.64B $\pm$ 3.83B & 5.17B $\pm$ 5.32B & 0.553 $\pm$ 0.636 & 77.8\% $\pm$ 11.1\% \\
Filing-only & 336 & 0.212 $\pm$ 0.072 & 0.313 $\pm$ 0.170 & 9.12B $\pm$ 7.12B & 22.49B $\pm$ 20.14B & -15.015 $\pm$ 29.041 & 64.8\% $\pm$ 9.0\% \\
MD\&A-only & 336 & 0.229 $\pm$ 0.184 & 0.475 $\pm$ 0.677 & 7.98B $\pm$ 8.04B & 14.07B $\pm$ 13.22B & -63.910 $\pm$ 180.410 & 74.4\% $\pm$ 6.6\% \\
\midrule
\rowcolor{gray!8}\multicolumn{8}{l}{\emph{History-Only Time-Series Models}} \\
TimeGPT-1 & 336 & 0.097 $\pm$ 0.047 & 0.097 $\pm$ 0.047 & 2.94B $\pm$ 3.23B & 3.75B $\pm$ 4.05B & 0.842 $\pm$ 0.128 & 71.0\% $\pm$ 15.2\% \\
Chronos-Bolt Base (205M) & 336 & 0.096 $\pm$ 0.045 & 0.095 $\pm$ 0.046 & 2.95B $\pm$ 3.27B & 3.73B $\pm$ 4.06B & 0.843 $\pm$ 0.114 & 69.1\% $\pm$ 16.4\% \\
\midrule
\rowcolor{gray!8}\multicolumn{8}{l}{\emph{Direct Same-Task LLM Baselines}} \\
History-only & 336 & 0.098 $\pm$ 0.057 & 0.094 $\pm$ 0.056 & 2.35B $\pm$ 2.13B & 3.27B $\pm$ 2.99B & 0.826 $\pm$ 0.195 & 76.2\% $\pm$ 10.8\% \\
History + Guidance & 336 & \underline{0.066 $\pm$ 0.033} & \underline{0.063 $\pm$ 0.030} & \underline{2.07B $\pm$ 2.07B} & \underline{2.75B $\pm$ 2.81B} & \underline{0.906 $\pm$ 0.093} & 81.2\% $\pm$ 11.4\% \\
\rowcolor{blue!5} History + Transcript & 336 & 0.067 $\pm$ 0.044 & 0.090 $\pm$ 0.120 & 2.55B $\pm$ 2.83B & 5.26B $\pm$ 6.88B & 0.295 $\pm$ 1.600 & \textbf{85.5\% $\pm$ 11.2\%} \\
\midrule
\rowcolor{gray!8}\multicolumn{8}{l}{\emph{Anchor-Relative Forecasting}} \\
Statistical Anchor (matched control) & 336 & 0.070 $\pm$ 0.036 & 0.067 $\pm$ 0.033 & 2.25B $\pm$ 2.33B & 2.96B $\pm$ 3.10B & 0.903 $\pm$ 0.087 & 74.4\% $\pm$ 20.9\% \\
\rowcolor{yellow!8} CAME (ours) & 336 & \textbf{0.059 $\pm$ 0.035} & \textbf{0.058 $\pm$ 0.035} & \textbf{1.81B $\pm$ 1.84B} & \textbf{2.39B $\pm$ 2.36B} & \textbf{0.929 $\pm$ 0.067} & \underline{84.0\% $\pm$ 11.0\%} \\
\midrule
\rowcolor{gray!8}\multicolumn{8}{l}{\emph{CAME Improvement}} \\
\rowcolor{yellow!8} CAME vs Statistical Anchor & 336 & +15.3\% & +13.1\% & +19.6\% & +19.3\% & +0.026 & +9.6 pp \\
\rowcolor{yellow!8} CAME vs History + Guidance & 336 & +9.6\% & +7.6\% & +12.6\% & +13.3\% & +0.023 & +2.8 pp \\
\bottomrule
\end{tabular}%
}
\caption{\textbf{Overall Performance on the Development-Inclusive 336-Row Rolling Backtest.} Metric rows are company-level mean $\pm$ sample standard deviation; $N$ is row coverage. MAE/RMSE use company reporting-currency billions (USD except ASML in EUR), so their cross-company aggregates are descriptive rather than single-currency totals. Lower errors and higher $R^2$/DA are better. Improvement rows report relative reference-minus-CAME error, absolute $\Delta R^2=R^2_{\mathrm{CAME}}-R^2_{\mathrm{reference}}$, and DA percentage-point changes; positive values favor CAME.}
\label{tab:main}
\end{table*}

\noindent \textbf{Evaluation Metrics.} Headline metric is company-equal macro sMAPE, normalizing revenue scale before averaging across companies. Table~\ref{tab:main} computes each metric within company, then reports the mean $\pm$ sample standard deviation across companies; $N$ reports the number of company-quarter observations and is not averaged. Because every company contributes 28 rows, the mean company MAE equals pooled MAE on this evaluation set. Appendix~\ref{app:metric-details} defines all six metrics, including the zero-variance and first-row rules. Paired-bootstrap intervals are computed over company-level macro-sMAPE deltas, with reference minus CAME so positive values favor CAME.

\noindent \textbf{Implementation Details.}
Non-LLM components (anchor selection, residual scoring, validation, evaluation) were executed locally. Closed-model LLM calls used the official OpenAI API for taxonomy discovery, evidence relation/event extraction, and all main-table direct LLM comparators. Evidence event extraction and the direct comparators used temperature 0.0; relation extraction omitted temperature and therefore used the provider default. Supplementary open-weight models were deployed locally using \texttt{vLLM} or Ollama endpoints on a single NVIDIA RTX A5000 (24GB) GPU. Refer to Appendix~\ref{app:prompt-comparator-details} for model identifiers, schema constraints, and comprehensive hardware specifications.

\subsection{Overall Performance}

CAME achieves the lowest sMAPE, MAPE, MAE, and RMSE and the highest $R^2$ among the reported methods on the development-inclusive 336-row rolling-backtest surface; History + Transcript retains the highest DA. Relative to the matched Statistical Anchor, it reduces macro sMAPE by 15.3\% and MAE by 19.6\%, while DA rises from 74.4\% to 84.0\%, an absolute gain of 9.6 percentage points. Against the direct same-task History + Guidance baseline, CAME improves sMAPE by 9.6\%, MAPE by 7.6\%, MAE by 12.6\%, and RMSE by 13.3\%, while raising $R^2$ by 0.023 and DA by 2.8 percentage points.

These results should be interpreted as aggregate improvement rather than universal per-company dominance. CAME improves over the matched anchor for all 12 companies by sMAPE and has lower company-level sMAPE than History + Guidance for nine of 12; AVGO, GOOGL, and TSLA favor the direct comparator. ``History + Transcript,'' which conditions on revenue history and full earnings-call transcripts, achieves the strongest directional accuracy but remains weaker than CAME on revenue-level error. Appendix Table~\ref{tab:locked-consistency} reports the 96-row temporal held-out evaluation and its initial-firm and company-held-out partitions.

\begin{table*}[!ht]
\centering
\scriptsize
\setlength{\tabcolsep}{4pt}
\begin{tabular}{lcrrr}
\toprule
\multicolumn{5}{l}{\textbf{Panel A. Company-equal metrics}} \\
\midrule
\textbf{Guidance Slice} & \textbf{N} & \textbf{sMAPE} & \textbf{MAPE} & \textbf{$R^2$} \\
\midrule
Explicit Numeric & 165 & $.0412/.0369\,(+.0043)$ & $.0408/.0369\,(+.0039)$ & $.9468/.9655\,(+.0187)$ \\
Non-Explicit & 65 & $.0994/.0981\,(+.0013)$ & $.0983/.0977\,(+.0006)$ & $-3.2947/-3.4395\,(-.1449)$ \\
No Guidance & 106 & $.1442/.0857\,(+.0585)$ & $.1310/.0814\,(+.0497)$ & $.4180/.7342\,(+.3162)$ \\
\bottomrule
\end{tabular}

\vspace{3pt}

\begin{tabular}{lcrrr}
\toprule
\multicolumn{5}{l}{\textbf{Panel B. Business-scale error and direction}} \\
\midrule
\textbf{Guidance Slice} & \textbf{N} & \textbf{MAE (B)} & \textbf{RMSE (B)} & \textbf{DA} \\
\midrule
Explicit Numeric & 165 & $1.295/.952\,(+.343)$ & $1.927/1.522\,(+.405)$ & $88.5\%/91.7\%\,(+3.2\,\mathrm{pp})$ \\
Non-Explicit & 65 & $2.717/2.566\,(+.151)$ & $4.229/4.163\,(+.067)$ & $84.5\%/87.9\%\,(+3.4\,\mathrm{pp})$ \\
No Guidance & 106 & $3.453/2.683\,(+.770)$ & $4.476/3.133\,(+1.342)$ & $60.0\%/77.9\%\,(+17.9\,\mathrm{pp})$ \\
\bottomrule
\end{tabular}
\caption{\textbf{Guidance Availability on the Development-Inclusive 336-Row Backtest.} Cells show Anchor/CAME (delta): $\Delta=\mathrm{Anchor}-\mathrm{CAME}$ for errors and $\Delta=\mathrm{CAME}-\mathrm{Anchor}$ for $R^2$/DA, so positive favors CAME. MAE/RMSE use company reporting-currency billions (USD except ASML in EUR). sMAPE, MAPE, RMSE, and $R^2$ are company-equal (zero-variance slices omitted for $R^2$); MAE and DA are pooled. DA uses the previous same-slice actual and excludes each company's first slice row.}
\label{tab:guidance}
\end{table*}

\begin{table*}[!ht]
\centering
\footnotesize
\setlength{\tabcolsep}{2.5pt}
\begin{tabular}{lrrrrrrr}
\toprule
\textbf{Variant} & \textbf{Full} $\downarrow$ & \textbf{Held-out (96)} $\downarrow$ & \textbf{Explicit} $\downarrow$ & \textbf{Non-exp.} $\downarrow$ & \textbf{No-guid.} $\downarrow$ & \textbf{F.MAE} $\downarrow$ & \textbf{DA} $\uparrow$ \\
\midrule
\rowcolor{yellow!8} CAME & \textbf{0.0593} & \textbf{0.0378} & \textbf{0.0369} & 0.0981 & \textbf{0.0857} & \textbf{1.81B} & \textbf{84.0\%} \\
w/o CurrentEvidence+TemporalMemory & 0.0683 & 0.0519 & 0.0381 & 0.0994 & 0.1276 & 2.25B & 79.9\% \\
w/o CurrentEvidence & 0.0671 & 0.0463 & 0.0375 & 0.1011 & 0.1267 & 2.16B & 79.9\% \\
w/o TemporalMemory & 0.0659 & 0.0458 & 0.0380 & \textbf{0.0954} & 0.1246 & 2.10B & 80.6\% \\
w/o AnchorMemory & 0.0654 & 0.0405 & 0.0370 & 0.0981 & 0.1379 & 1.98B & 81.8\% \\
w/o ExplicitGuidance & 0.0616 & 0.0386 & 0.0408 & 0.0981 & \textbf{0.0857} & 2.00B & 82.7\% \\
\bottomrule
\end{tabular}%
\caption{\textbf{Ablation Study.} Full and Held-out denote the development-inclusive 336-row and temporal held-out 96-row evaluations. The first five metric columns are company-equal macro sMAPE; F.MAE and DA are 336-row mean company metrics, with F.MAE in company reporting-currency billions. Each removal reruns the reduced path and may recompute dependent proposals and weights, so contrasts are interaction-inclusive end-to-end ablations, not isolated or additive effects. Lower error and higher DA are better.}
\label{tab:ablation}
\end{table*}

\subsection{Guidance-Availability Analysis}

\noindent \textbf{Setup.}
To rigorously analyze CAME's behavior, we partition forecast instances by forecast-time guidance availability: ``Explicit Numeric Guidance'' (numeric total-revenue range or midpoint), ``Non-Explicit Guidance'' (qualitative or indirect forward-looking guidance), and ``No Guidance'' (no target-quarter total-revenue guidance). CAME is designed so anchor memory does not override guidance-bearing rows. This analysis therefore tests whether guidance-gated anchor memory acts as a targeted correction for no-guidance cases, rather than as an implicit mechanism for rewriting management guidance. This distinction is critical, as explicit and non-explicit guidance rows already contain management's information; an indiscriminate anchor-memory application risks complicating interpretation and double-counting.

\noindent \textbf{Results.}
Table~\ref{tab:guidance} quantitatively reflects this design boundary. CAME's largest gains occur in the ``No Guidance'' slice, where all six reported point estimates favor CAME. ``Explicit Numeric Guidance'' gains are smaller, whereas ``Non-Explicit Guidance'' is mixed: four error metrics and DA improve, while the macro-$R^2$ decline is dominated by INTC's two-row slice under company-equal averaging. Both guidance-bearing sMAPE intervals include zero. The robust interpretation is therefore not across-the-board dominance, but rather that CAME's strongest and most policy-relevant effect is to correct Statistical Anchor weaknesses when guidance is unavailable.

\subsection{Ablation Study}

\noindent \textbf{Setup.}
Table~\ref{tab:ablation} reports end-to-end component removals on the development-inclusive 336-row and temporal held-out 96-row evaluations. Each removal reruns the reduced path and may recompute dependent proposals and weights, so contrasts are interaction-inclusive and non-additive rather than isolated zeroing effects. The full CAME path achieves the best full-surface sMAPE, temporal held-out sMAPE, Full MAE, and Direction Accuracy. Moreover, every component removal worsens full-surface sMAPE, supporting the unified component path rather than a hidden per-slice selector. The slice-level results further show scoped-component behavior: temporal memory is not uniformly dominant, and the removal of guidance has no effect on no-guidance rows where the expert is inactive.

\noindent \textbf{Component Roles.}
Removing Anchor Memory produces the largest no-guidance degradation (sMAPE $0.085681\!\rightarrow\!0.137930$); removing Current Evidence and Temporal Memory raises full sMAPE from 0.059253 to 0.068279 and lowers DA from 84.0\% to 79.9\%. Removing Explicit Guidance worsens its eligible Explicit slice and aggregate metrics while leaving the other slices unchanged.

\noindent \textbf{Sensitivity and Boundaries.}
Eight one-factor memory/retrieval alternatives keep full sMAPE within 0.059180--0.059557 and all anchor margins positive. Two paired extraction repeats favor a broader rule on the full surface by 0.000524 on average but disfavor it on held-out data by 0.000487; these are directional checks, and the retained settings remain unchanged. Removing Temporal Memory slightly improves the Non-exp. slice, consistent with the overlap between current and historical signals. Under 5\%/10\% polarity, omission, and false-positive-horizon stress, full and held-out anchor margins remain positive, but one 10\% seed per stress type loses on Non-exp. Guardrail attenuation of 65.1--77.1\% measures prediction movement, not a guard-disabled comparison, sMAPE, or real-world prevention probability (Appendix~\ref{app:sensitivity-validation}).

\subsection{Paired Bootstrap Uncertainty}

The paired-bootstrap analysis supports the headline and temporal held-out Statistical Anchor contrasts: the development-inclusive gain remains positive (+0.010697, 95\% CI $[+0.007463,+0.014067]$), and the temporal held-out gain is strictly positive (+0.009360, 95\% CI $[+0.004235,+0.016219]$). The company-held-out temporal and No Guidance intervals are likewise positive, although No Guidance is a development-inclusive subgroup; the two guidance-bearing intervals touch or cross zero. Against the direct LLM baselines, both development-inclusive intervals and temporal held-out History-only are positive, whereas temporal held-out History + Guidance crosses zero. Appendix~\ref{app:rq4-bootstrap-details} reports the complete uncertainty results.

\subsection{Forecast Traceability}
\label{sec:forecast-traceability}
CAME reports the selected Statistical Anchor, memory correction, support/conflict cards, temporal matches, and final prediction; this is provenance, not an accuracy guarantee. In an outcome-blind 120-card evaluation, source support was full for 102 and partial for 18, target-company revenue validity held for 96/120, target-quarter support was strict for 55 and partial for 41, and polarity matched consensus for 86/110 evaluable cards with 10 strict reversals. These sampled semantic judgments do not establish causal correctness or trace utility. Appendix Figure~\ref{fig:app-aapl-trace-audit} shows an AAPL case in which positive product and Services evidence, moderated by an FX headwind, moves the forecast toward the realized target. Figure~\ref{fig:app-memory-boundary-cases} shows the complementary TSLA failure boundary: a high anchor, insufficient downward evidence, and positive temporal analogs leave the forecast above the realized target. Retrospective cards are post-hoc only and never enter forecasting.

\section{Conclusion}
We presented CAME, a company-aware residual framework for quarter-ahead revenue forecasting. On a development-inclusive 12-company, 336-instance benchmark covering large technology and platform firms, CAME lowers aggregate revenue error over the matched Statistical Anchor, with the strongest within-panel evidence when explicit guidance is unavailable. Against History + Guidance, CAME leads all six development-inclusive aggregate metrics and all four error metrics plus DA in the 96-row temporal held-out evaluation; the direct comparator has a higher held-out $R^2$. The results support anchor-relative, temporally eligible, source-linked refinement within the evaluated panel.

\section*{Limitations}
\paragraph{Scope of Empirical Gains.}
CAME is evaluated on 12 large public technology and platform companies with relatively complete reporting histories. Although it improves aggregate revenue-level error over the matched Statistical Anchor, gains remain heterogeneous across companies, metrics, and guidance slices, and History + Guidance remains stronger for some firms. Broader sectors, smaller firms, sparse-disclosure settings, and frequent segment redefinitions remain important future tests. CAME's traces provide provenance for forecast adjustments but do not establish causal correctness or validated trace utility.

\paragraph{Comparator and Operational Limits.}
Some comparator rows, including filing-only and MD\&A-only settings, are adapted document-only controls rather than equally strong source-policy baselines. Identity masking removes only company-name tokens and does not prevent re-identification from retained financial context. CAME also requires nontrivial preprocessing, schema maintenance, and quality assurance. The reported operational-cost estimate excludes source acquisition, schema setup, human quality assurance, comparator execution, historical backfill, and unobserved retries, and is therefore not an end-to-end deployment estimate.

\paragraph{Human Evaluation and Risks.}
The evidence-card audit was conducted by two authors and one colleague from the same laboratory, which may introduce evaluator bias; the sampled judgments do not establish population-level extraction validity, causal adjustment correctness, or trace utility. Forecasts may also be misused as investment advice or trading signals and may fail under incomplete disclosures, extraction errors, or LLM reasoning errors. We report research results rather than financial recommendations.

\section*{Ethical Considerations}\label{Ethics}
The study uses publicly available corporate disclosures and revenue records. Because public access does not imply redistribution rights, release is limited to author-controlled code and author-created, quote-free normalized card fields, trace fields, and provenance metadata; private verbatim \texttt{source\_span} values, other quote-bearing artifacts, and raw third-party corpora remain excluded without confirmed permission.

The internal evidence-card audit used three human annotators: two authors and one colleague from the same laboratory. The reported claims concern the research artifacts rather than annotator characteristics. We therefore report a bounded internal artifact audit rather than independent human validation.

\section*{Acknowledgments}
This work was supported by the National Science and Technology Council, Taiwan, under Grant NSTC-114-2639-E-A49-001-ASP and NSTC-115-2221-E-A49-120-MY3. We thank the anonymous reviewers and the meta-reviewer for their constructive feedback.

ChatGPT (OpenAI) was used to assist with language editing, code development and debugging, and the organization and checking of experimental results. The authors independently determined the research design, methodology, and interpretation of the results, verified all numerical results and citations, and are fully responsible for the final manuscript.

\bibliography{custom}

@inproceedings{koval-etal-2024-financial,
    title = "Financial Forecasting from Textual and Tabular Time Series",
    author = "Koval, Ross  and
      Andrews, Nicholas  and
      Yan, Xifeng",
    booktitle = "Findings of the Association for Computational Linguistics: EMNLP",
    month = nov,
    year = "2024",
    pages = "8289--8300"
}

@inproceedings{koval-etal-2024-learning,
    title = "Learning to Compare Financial Reports for Financial Forecasting",
    author = "Koval, Ross  and
      Andrews, Nicholas  and
      Yan, Xifeng",
    booktitle = "Findings of the Association for Computational Linguistics: EACL 2024",
    year = "2024",
    pages = "500--512"
}

@article{divo2025forecasting,
  title={Forecasting Company Fundamentals},
  author={Felix Divo and Eric Endress and Kevin Endler and Kristian Kersting and Devendra Singh Dhami},
  journal={Transactions on Machine Learning Research (TMLR)},
  issn={2835-8856},
  year={2025}
}

@inproceedings{sawhney-etal-2020-voltage,
    title = "{V}ol{TAGE}: Volatility Forecasting via Text Audio Fusion with Graph Convolution Networks for Earnings Calls",
    author = "Sawhney, Ramit  and
      Khanna, Piyush  and
      Aggarwal, Arshiya  and
      Jain, Taru  and
      Mathur, Puneet  and
      Shah, Rajiv Ratn",
    booktitle = "Proceedings of the Conference on Empirical Methods in Natural Language Processing (EMNLP)",
    year = "2020",
    pages = "8001--8013"
}

@inproceedings{zhu-etal-2021-tat,
    title = "{TAT}-{QA}: A Question Answering Benchmark on a Hybrid of Tabular and Textual Content in Finance",
    author = "Zhu, Fengbin  and
      Lei, Wenqiang  and
      Huang, Youcheng  and
      Wang, Chao  and
      Zhang, Shuo  and
      Lv, Jiancheng  and
      Feng, Fuli  and
      Chua, Tat-Seng",
    booktitle = "Proceedings of the Annual Meeting of the Association for Computational Linguistics (ACL)",
    year = "2021",
    pages = "3277--3287"
}

@inproceedings{chen-etal-2021-finqa,
    title = "{F}in{QA}: A Dataset of Numerical Reasoning over Financial Data",
    author = "Chen, Zhiyu  and
      Chen, Wenhu  and
      Smiley, Charese  and
      Shah, Sameena  and
      Borova, Iana  and
      Langdon, Dylan  and
      Moussa, Reema  and
      Beane, Matt  and
      Huang, Ting-Hao  and
      Routledge, Bryan  and
      Wang, William Yang",
    booktitle = "Proceedings of the Conference on Empirical Methods in Natural Language Processing (EMNLP)",
    year = "2021",
    pages = "3697--3711"
}

@inproceedings{chen-etal-2022-convfinqa,
    title = "{C}onv{F}in{QA}: Exploring the Chain of Numerical Reasoning in Conversational Finance Question Answering",
    author = "Chen, Zhiyu  and
      Li, Shiyang  and
      Smiley, Charese  and
      Ma, Zhiqiang  and
      Shah, Sameena  and
      Wang, William Yang",
    booktitle = "Proceedings of the Conference on Empirical Methods in Natural Language Processing (EMNLP)",
    year = "2022",
    pages = "6279--6292"
}

@inproceedings{reddy-etal-2024-docfinqa,
    title = "{D}oc{F}in{QA}: A Long-Context Financial Reasoning Dataset",
    author = "Reddy, Varshini  and
      Koncel-Kedziorski, Rik  and
      Lai, Viet Dac  and
      Krumdick, Michael  and
      Lovering, Charles  and
      Tanner, Chris",
    booktitle = "Proceedings of the Annual Meeting of the Association for Computational Linguistics (ACL)",
    year = "2024",
    pages = "445--458"
}

@article{LIM20211748,
  title = {Temporal Fusion Transformers for interpretable multi-horizon time series forecasting},
  journal = {International Journal of Forecasting},
  volume = {37},
  number = {4},
  pages = {1748-1764},
  year = {2021},
  author = {Bryan Lim and Sercan Ö. Arık and Nicolas Loeff and Tomas Pfister}
}

@inproceedings{nie2023a,
  title={A Time Series is Worth 64 Words:  Long-term Forecasting with Transformers},
  author={Yuqi Nie and Nam H Nguyen and Phanwadee Sinthong and Jayant Kalagnanam},
  booktitle={The International Conference on Learning Representations (ICLR)},
  year={2023},
}

@inproceedings{jin2024timellm,
  title={Time-{LLM}: Time Series Forecasting by Reprogramming Large Language Models},
  author={Ming Jin and Shiyu Wang and Lintao Ma and Zhixuan Chu and James Y. Zhang and Xiaoming Shi and Pin-Yu Chen and Yuxuan Liang and Yuan-Fang Li and Shirui Pan and Qingsong Wen},
  booktitle={The International Conference on Learning Representations (ICLR)},
  year={2024},
}

@inproceedings{li-lu-2026-decoding,
    title = "Decoding the Market{'}s Pulse: Context-Enriched Agentic Retrieval Augmented Generation for Predicting Post-Earnings Price Shocks",
    author = "Li, Chenhui  and
      Lu, Weihai",
    booktitle = "Proceedings of the Conference of the {E}uropean Chapter of the {A}ssociation for {C}omputational {L}inguistics (EACL)",
    year = "2026",
    pages = "3055--3073",
}

@inproceedings{10.5555/3495724.3496517,
  author = {Lewis, Patrick and Perez, Ethan and Piktus, Aleksandra and Petroni, Fabio and Karpukhin, Vladimir and Goyal, Naman and K\"{u}ttler, Heinrich and Lewis, Mike and Yih, Wen-tau and Rockt\"{a}schel, Tim and Riedel, Sebastian and Kiela, Douwe},
  title = {Retrieval-augmented generation for knowledge-intensive NLP tasks},
  year = {2020},
  booktitle = {Proceedings of the International Conference on Neural Information Processing Systems (NeurIPS)}
}

@article{10.5555/196108.196115,
  author = {Aamodt, Agnar and Plaza, Enric},
  title = {Case-based reasoning: foundational issues, methodological variations, and system approaches},
  year = {1994},
  issue_date = {March 1994},
  publisher = {IOS Press},
  address = {NLD},
  volume = {7},
  number = {1},
  issn = {0921-7126},
  journal = {AI Commun.},
  month = mar,
  pages = {39–59},
  numpages = {21},
  doi = {10.3233/AIC-1994-7104}
}

@inproceedings{koval-etal-2023-forecasting,
    title = "Forecasting Earnings Surprises from Conference Call Transcripts",
    author = "Koval, Ross  and
      Andrews, Nicholas  and
      Yan, Xifeng",
    booktitle = "Findings of the Association for Computational Linguistics: ACL",
    year = "2023",
    pages = "8197--8209"
}

@inproceedings{qin-yang-2019-say,
    title = "What You Say and How You Say It Matters: Predicting Stock Volatility Using Verbal and Vocal Cues",
    author = "Qin, Yu  and
      Yang, Yi",
    booktitle = "Proceedings of the Annual Meeting of the Association for Computational Linguistics (ACL)",
    year = "2019",
    pages = "390--401"
}

@misc{araci2020finbert,
  title={{FinBERT}: Financial Sentiment Analysis with Pre-trained Language Models},
  author={Dogu Araci and Zulkuf Genc},
  year={2020},
  url={https://openreview.net/forum?id=HylznxrYDr}
}

@article{loughran2011liability,
  title={When is a liability not a liability? Textual analysis, dictionaries, and 10-Ks},
  author={Loughran, Tim and McDonald, Bill},
  journal={The Journal of finance},
  volume={66},
  number={1},
  pages={35--65},
  year={2011}
}

@article{shah2023zero,
  title={Zero is not hero yet: Benchmarking zero-shot performance of llms for financial tasks},
  author={Shah, Agam and Chava, Sudheer},
  journal={arXiv preprint arXiv:2305.16633},
  year={2023}
}

@inproceedings{ni2024harnessing,
  title={Harnessing earnings reports for stock predictions: A qlora-enhanced llm approach},
  author={Ni, Haowei and Meng, Shuchen and Chen, Xupeng and Zhao, Ziqing and Chen, Andi and Li, Panfeng and Zhang, Shiyao and Yin, Qifu and Wang, Yuanqing and Chan, Yuxi},
  booktitle={The International conference on data-driven optimization of complex systems (DOCS)},
  pages={909--915},
  year={2024}
}

@article{hyndman2008automatic,
  title={Automatic time series forecasting: the forecast package for R},
  author={Hyndman, Rob J and Khandakar, Yeasmin},
  journal={Journal of statistical software},
  volume={27},
  pages={1--22},
  year={2008}
}

@article{HYNDMAN2002439,
  title = {A state space framework for automatic forecasting using exponential smoothing methods},
  journal = {International Journal of Forecasting},
  volume = {18},
  number = {3},
  pages = {439-454},
  year = {2002},
  author = {Rob J Hyndman and Anne B Koehler and Ralph D Snyder and Simone Grose}
}

@article{hoerl1970ridge,
  title={Ridge regression: Biased estimation for nonorthogonal problems},
  author={Hoerl, Arthur E and Kennard, Robert W},
  journal={Technometrics},
  volume={12},
  number={1},
  pages={55--67},
  year={1970}
}

@article{zou2005regularization,
  title={Regularization and variable selection via the elastic net},
  author={Zou, Hui and Hastie, Trevor},
  journal={Journal of the Royal Statistical Society Series B: Statistical Methodology},
  volume={67},
  number={2},
  pages={301--320},
  year={2005}
}

@article{breiman2001random,
  author = {Breiman, Leo},
  title = {Random Forests},
  year = {2001},
  publisher = {Kluwer Academic Publishers},
  address = {USA},
  volume = {45},
  number = {1},
  journal = {Machine Learning},
  month = oct,
  pages = {5–32}
}

@inproceedings{10.1145/2939672.2939785,
  author = {Chen, Tianqi and Guestrin, Carlos},
  title = {{XGBoost}: A Scalable Tree Boosting System},
  year = {2016},
  booktitle = {Proceedings of the ACM International Conference on Knowledge Discovery and Data Mining (SIGKDD)},
  pages = {785–794}
}

@inproceedings{10.5555/3294996.3295074,
  author = {Ke, Guolin and Meng, Qi and Finley, Thomas and Wang, Taifeng and Chen, Wei and Ma, Weidong and Ye, Qiwei and Liu, Tie-Yan},
  title = {LightGBM: a highly efficient gradient boosting decision tree},
  year = {2017},
  booktitle = {Proceedings of the International Conference on Neural Information Processing Systems (NeurIPS)},
  pages = {3149–3157}
}

@misc{mistralai2025small3,
  title = {{Mistral Small 3}},
  author = {{Mistral AI Team}},
  year = {2025},
  url = {https://mistral.ai/news/mistral-small-3/},
}

@article{grattafiori2024llama3herd,
  title={The llama 3 herd of models},
  author={Grattafiori, Aaron and Dubey, Abhimanyu and Jauhri, Abhinav and Pandey, Abhinav and Kadian, Abhishek and Al-Dahle, Ahmad and Letman, Aiesha and Mathur, Akhil and Schelten, Alan and Vaughan, Alex and others},
  journal={arXiv preprint arXiv:2407.21783},
  year={2024}
}

@article{Yang2024Qwen25TR,
  title={{Qwen2.5} Technical Report},
  author={{Qwen Team}},
  journal={ArXiv},
  year={2024},
  volume={abs/2412.15115}
}

@article{yang2025qwen3,
  title={{Qwen3} technical report},
  author={Yang, An and Li, Anfeng and Yang, Baosong and Zhang, Beichen and Hui, Binyuan and Zheng, Bo and Yu, Bowen and Gao, Chang and Huang, Chengen and Lv, Chenxu and others},
  journal={arXiv preprint arXiv:2505.09388},
  year={2025}
}

@article{seabold2010,
  author = {Seabold, Skipper and Perktold, Josef},
  title = {Statsmodels: Econometric and Statistical Modeling with Python},
  journal = {SciPy 2010},
  year = {2010}
}

@misc{garza2024timegpt,
      title={{TimeGPT}-1}, 
      author={Azul Garza and Cristian Challu and Max Mergenthaler-Canseco},
      year={2024},
      eprint={2310.03589},
      archivePrefix={arXiv},
      primaryClass={cs.LG},
      url={https://arxiv.org/abs/2310.03589}, 
}

@article{ansari2024chronos,
  title={Chronos: Learning the Language of Time Series},
  author={Ansari, Abdul Fatir and Stella, Lorenzo and Turkmen, Caner and Zhang, Xiyuan and Mercado, Pedro and Shen, Huibin and Shchur, Oleksandr and Rangapuram, Syama Sundar and Pineda Arango, Sebastian and Kapoor, Shubham and Zschiegner, Jasper and Maddix, Danielle C. and Wang, Hao and Mahoney, Michael W. and Torkkola, Kari and Wilson, Andrew Gordon and Bohlke-Schneider, Michael and Wang, Yuyang},
  journal={Transactions on Machine Learning Research},
  issn={2835-8856},
  year={2024},
  url={https://openreview.net/forum?id=gerNCVqqtR}
}

@misc{openai2025gpt41,
  title = {Introducing {GPT-4.1} in the {API}},
  author = {{OpenAI}},
  year = {2025},
  month = apr,
  url = {https://openai.com/index/gpt-4-1/},
}

@misc{openai2024gpt4omini,
  title = {{GPT-4o mini}: advancing cost-efficient intelligence},
  author = {{OpenAI}},
  year = {2024},
  month = jul,
  url = {https://openai.com/index/gpt-4o-mini-advancing-cost-efficient-intelligence/},
}

\clearpage
\appendix

\setcounter{table}{0}
\renewcommand{\thetable}{A\arabic{table}}


\section{Reproducibility Statement}\label{Reproducibility_statement}
Code, quote-free normalized card fields, trace fields, provenance metadata, and predictions will be released at \url{https://github.com/yy1200/came-revenue-forecasting}, including shared settings and schemas, executable History-only and History + Guidance prompt specifications, split and anchor rules, validation and evaluation scripts, all 336 CAME predictions, and source- and redistribution-provenance labels. The included deterministic replay procedure reproduces 336/336 forecasts within absolute and relative tolerances of $10^{-6}$ and $10^{-12}$ and recomputes the development-inclusive and temporal held-out metrics without institution-exclusive access. It does not reproduce source acquisition, document extraction, or source-grounding judgments; private verbatim \texttt{source\_span} values, raw third-party corpora, and direct-comparator API responses and cached outputs are excluded.

\section{Error Analysis}
We illustrate three error modes observed in the diagnostic analysis.

\paragraph{Redundant Memory Signals.}
The clearest boundary appears in the non-explicit-guidance slice, where removing Temporal Memory lowers sMAPE from 0.0981 to 0.0954 even though it worsens aggregate and temporal held-out performance. Qualitative forward-looking commentary can overlap with current evidence or the Statistical Anchor; retrieved analogs then add directional reinforcement but limited independent magnitude calibration, so the correction can be too large even when it moves in the correct direction.

\paragraph{Anchor Inertia under Sharp Shifts.}
Some large errors arise when the Statistical Anchor begins materially high or low relative to a quarter with a sharper transition. Although CAME can adjust the anchor, the residual correction may remain too weak when forecast-time evidence is mixed or does not provide a sufficiently strong counter-signal. Figure~\ref{fig:app-memory-boundary-cases} illustrates this boundary: positive evidence-memory signals and a high initial anchor leave the guarded forecast above the realized target.

\paragraph{Weak Analog Coverage.}
A further limitation appears when the target quarter includes firm-specific events with only weak historical analogs, such as product-transition timing, monetization changes, or segment-specific supply constraints. In these cases, the system may retrieve relevant evidence yet still misestimate effect size because the eligible memory bank provides only partial precedent for the magnitude of the revenue impact. This is reflected in the unmatched or retrospective factors highlighted in Figures~\ref{fig:app-aapl-trace-audit} and~\ref{fig:app-memory-boundary-cases}.
Taken together, these examples highlight one important limitation: calibrating the magnitude of a correction when historical analogs are partial or redundant. They do not establish that missing relevant evidence is unimportant or that a particular precedent pattern is generally more reliable.

\section{Metric Definitions}
\label{app:metric-details}

For a row set $\mathcal{I}$ with $n=|\mathcal{I}|$, realized revenue $y_i>0$, forecast $\hat y_i$, and mean actual $\bar y$, we use
\begin{equation}
\begin{aligned}
\mathrm{sMAPE} &= \frac{1}{n}\sum_{i\in\mathcal I}
\frac{2|\hat y_i-y_i|}{|y_i|+|\hat y_i|},\\
\mathrm{MAPE} &= \frac{1}{n}\sum_{i\in\mathcal I}
\frac{|\hat y_i-y_i|}{|y_i|},\\
\mathrm{MAE} &= \frac{1}{n}\sum_{i\in\mathcal I}|\hat y_i-y_i|,\\
\mathrm{RMSE} &= \sqrt{\frac{1}{n}\sum_{i\in\mathcal I}(\hat y_i-y_i)^2},\\
R^2 &= 1-\frac{\sum_{i\in\mathcal I}(\hat y_i-y_i)^2}
{\sum_{i\in\mathcal I}(y_i-\bar y)^2}.
\end{aligned}
\end{equation}
Metrics labeled company-equal are computed separately within each eligible company and then averaged with equal company weight. Table~\ref{tab:main} also reports the sample standard deviation across companies. For guidance slices and Table~\ref{tab:locked-consistency}, sMAPE, MAPE, RMSE, and $R^2$ are company-equal, while MAE and DA are pooled over eligible rows. Table~\ref{tab:open-model-locked} uses pooled RMSE for its same-row open-weight comparison and labels that column explicitly. A company or company-slice with zero actual variance has undefined $R^2$ and is excluded only from the $R^2$ macro average; no zero value is imputed.

For surface or slice $S$, let $p_S(i)$ be the previous evaluated row for the same company within $S$. Define predicted and realized directions by
\begin{equation}
\begin{aligned}
d^{\mathrm{pred}}_{i,S} &= \mathrm{sgn}(\hat y_i-y_{p_S(i)}),\\
d^{\mathrm{actual}}_{i,S} &= \mathrm{sgn}(y_i-y_{p_S(i)}).
\end{aligned}
\end{equation}
Directional accuracy is then
\begin{equation}
\mathrm{DA}_S=\frac{1}{|\mathcal I_S^{\mathrm{dir}}|}
\sum_{i\in\mathcal I_S^{\mathrm{dir}}}
\mathbf 1[d^{\mathrm{pred}}_{i,S}=d^{\mathrm{actual}}_{i,S}],
\end{equation}
where $\mathcal I_S^{\mathrm{dir}}$ excludes each company's first row in $S$. Thus guidance-slice DA uses the previous same-slice actual and excludes the first row of each company-slice. Error deltas are reference minus CAME; $R^2$ and DA deltas are CAME minus reference. Positive deltas therefore favor CAME, and DA deltas are percentage points.

\section{Evaluation Sets and Roles}
\label{app:evaluation-surfaces}

Table~\ref{tab:evaluation-surfaces} distinguishes the development-inclusive 336-row rolling backtest from the FY2024 Q1--FY2025 Q4 temporal held-out window. The nine additional companies are held out for method and protocol selection; their FY2019--FY2023 rows provide historical context and online history for later company-held-out forecasts. Method and protocol selection used only AAPL, NVDA, and AVGO through FY2023 Q4. All settings were then fixed before evaluating the 96-row temporal held-out window and the nine-company-held-out robustness panel. Every firm uses the same settings without per-company tuning or method selection.

\begin{table*}[t]
\centering
\small
\resizebox{\textwidth}{!}{%
\begin{tabular}{llll}
\toprule
Evaluation set & Companies & Quarters & Role \\
\midrule
Initial development period & AAPL, NVDA, AVGO & FY2019 Q1--FY2023 Q4 & initial method development \\
Initial-firm temporal held-out & AAPL, NVDA, AVGO & FY2024 Q1--FY2025 Q4 & temporal held-out evaluation \\
Company-held-out rolling backtest & nine additional firms & FY2019 Q1--FY2025 Q4 & company-held-out robustness evaluation \\
Company-held-out temporal subset & nine additional firms & FY2024 Q1--FY2025 Q4 & joint company/temporal held-out evaluation \\
Development-inclusive rolling backtest & all 12 firms & FY2019 Q1--FY2025 Q4 & development-inclusive aggregate evaluation \\
\bottomrule
\end{tabular}%
}
\caption{Evaluation sets and their roles.}
\label{tab:evaluation-surfaces}
\end{table*}

Table~\ref{tab:locked-consistency} reports the development-inclusive and temporal held-out results, including the initial-firm and company-held-out temporal partitions. TimeGPT-1 and Chronos-Bolt use only prior revenue history; the remaining rows report the Statistical Anchor, History + Guidance, and CAME. Metric aggregation and reporting-currency units follow Appendix~\ref{app:metric-details}.

\begin{table*}[t]
\centering
\scriptsize
\resizebox{\textwidth}{!}{%
\begin{tabular}{llrrrrrrr}
\toprule
Surface & Method & $N$ & sMAPE $\downarrow$ & MAPE $\downarrow$ & MAE$_{\mathrm{pool}}$ $\downarrow$ & RMSE$_{\mathrm{macro}}$ $\downarrow$ & $R^2$ $\uparrow$ & DA $\uparrow$ \\
\midrule
Development-inclusive rolling backtest & TimeGPT-1 & 336 & 0.097 & 0.097 & 2.94B & 3.75B & 0.842 & 71.0\% \\
& Chronos-Bolt Base & 336 & 0.096 & 0.095 & 2.95B & 3.73B & 0.843 & 69.1\% \\
& Statistical Anchor & 336 & 0.070 & 0.067 & 2.25B & 2.96B & 0.903 & 74.4\% \\
& History + Guidance & 336 & 0.066 & 0.063 & 2.07B & 2.75B & 0.906 & 81.2\% \\
& CAME & 336 & \textbf{0.059} & \textbf{0.058} & \textbf{1.81B} & \textbf{2.39B} & \textbf{0.929} & \textbf{84.0\%} \\
\midrule
Temporal held-out & TimeGPT-1 & 96 & 0.096 & 0.092 & 3.57B & 4.26B & -0.042 & 76.2\% \\
& Chronos-Bolt Base & 96 & 0.090 & 0.088 & 3.25B & 3.69B & 0.093 & 75.0\% \\
& Statistical Anchor & 96 & 0.047 & 0.046 & 1.93B & 2.29B & 0.611 & 79.8\% \\
& History + Guidance & 96 & 0.052 & 0.051 & 1.74B & 1.99B & \textbf{0.715} & 84.5\% \\
& CAME & 96 & \textbf{0.038} & \textbf{0.038} & \textbf{1.48B} & \textbf{1.86B} & 0.632 & \textbf{92.9\%} \\
\midrule
Initial-firm temporal held-out & Statistical Anchor & 24 & 0.049 & 0.046 & 1.85B & 2.23B & 0.854 & 85.7\% \\
& History + Guidance & 24 & 0.064 & 0.060 & 1.93B & 2.23B & 0.872 & 90.5\% \\
& CAME & 24 & \textbf{0.041} & \textbf{0.040} & \textbf{1.68B} & \textbf{2.13B} & \textbf{0.872} & \textbf{100.0\%} \\
\midrule
Company-held-out temporal subset & Statistical Anchor & 72 & 0.047 & 0.046 & 1.96B & 2.31B & 0.530 & 77.8\% \\
& History + Guidance & 72 & 0.048 & 0.048 & 1.68B & 1.91B & \textbf{0.663} & 82.5\% \\
& CAME & 72 & \textbf{0.037} & \textbf{0.038} & \textbf{1.42B} & \textbf{1.77B} & 0.552 & \textbf{90.5\%} \\
\bottomrule
\end{tabular}%
}
\caption{Development-inclusive and temporal held-out results.}
\label{tab:locked-consistency}
\end{table*}

Table~\ref{tab:open-model-locked} reports supplementary direct-LLM results on the 96-row temporal held-out evaluation. GPT-4.1-mini follows the direct-baseline setup in Appendix~\ref{app:prompt-comparator-details}; the open-weight rows use different input preparation and are therefore supplementary small/open-weight comparators rather than input-matched comparisons. All rows use deterministic decoding; metric aggregation and reporting-currency units follow Appendix~\ref{app:metric-details}.

\begin{table*}[t]
\centering
\scriptsize
\resizebox{\textwidth}{!}{%
\begin{tabular}{lrrrrrrr}
\toprule
Model & $N$ & sMAPE $\downarrow$ & MAPE $\downarrow$ & MAE$_{\mathrm{pool}}$ $\downarrow$ & RMSE$_{\mathrm{pool}}$ $\downarrow$ & $R^2$ $\uparrow$ & DA $\uparrow$ \\
\midrule
\rowcolor{gray!8}\multicolumn{8}{l}{\emph{Proprietary reference}} \\
GPT-4.1-mini & 96 & 0.052072 & 0.050990 & 1.741B & 2.379B & 0.715 & 84.5\% \\
\midrule
\rowcolor{gray!8}\multicolumn{8}{l}{\emph{Open-weight models}} \\
Mistral Small 3 (24B) & 96 & 0.050930 & 0.050471 & 1.771B & 2.547B & 0.529 & 91.7\% \\
Llama 3.1 8B & 96 & 0.063200 & 0.064114 & 1.957B & 2.874B & 0.132 & 86.9\% \\
Qwen2.5 14B & 96 & 0.061538 & 0.059264 & 2.722B & 4.735B & 0.562 & 84.5\% \\
\bottomrule
\end{tabular}%
}
\caption{Supplementary History + Guidance results on the 96-row temporal held-out evaluation.}
\label{tab:open-model-locked}
\end{table*}

Qwen3-14B \citep{yang2025qwen3} is omitted because, under the same prompt and prespecified parser, its History + Guidance outputs included invalid responses and extreme finite numerical outliers. Outputs were not corrected post hoc, so no comparable metric row is reported.

\section{Methodological Details}
\label{app:method-details}

\subsection{Inference Details}
Algorithm~\ref{alg:came-inference} summarizes the online CAME inference path: using only information available by the forecast cutoff, it forms a guidance-gated base anchor and composes current-evidence, temporal-memory, and eligible explicit-guidance residuals. The guarded forecast is emitted with a source-linked trace, and realized revenue enters history only after observation.
Throughout the algorithm, $\mathcal{O}_i^{\le t}$ denotes information released no later than the forecast cutoff, i.e., $r_j\le t_i$.

\par\medskip
\noindent\begin{minipage}{\columnwidth}
\refstepcounter{camealgorithm}
\label{alg:came-inference}
\small
\noindent\textbf{Algorithm~\thecamealgorithm:} CAME inference for one company-quarter
\par\smallskip
\hrule
\vspace{0.4ex}
\textbf{Input:} target company $c_i$, forecast time $t_i$, target quarter $q_i$, observable information $\mathcal{O}_i$.\par
\textbf{Output:} revenue forecast $\hat{y}_i$ and forecast trace $T_i$.
\vspace{0.4ex}
\hrule
\vspace{0.4ex}
\setcounter{camealgline}{0}
\begin{list}{\arabic{camealgline}}{
  \usecounter{camealgline}
  \setlength{\leftmargin}{1.8em}
  \setlength{\labelwidth}{1.1em}
  \setlength{\labelsep}{0.5em}
  \setlength{\itemsep}{0pt}
  \setlength{\parsep}{0pt}
  \setlength{\topsep}{0pt}
}
\item ${\mathcal{O}_i^{\le t}} \gets \mathrm{TemporalFilter}({\mathcal{O}_i},t_i)$
\item $E_i \gets g_{\mathrm{ext}}({\mathcal{N}_i^{\le t}}, c_i, q_i)$
\item $E_i \gets \mathrm{ValidateCards}(E_i)$ \hfill \textit{schema, source spans, quarter eligibility}
\item $b_i \gets \mathrm{StatisticalAnchor}(c_i,q_i,{\mathcal{O}_i^{\le t}})$
\item $\Delta_i^{mem} \gets 0$
\item \textbf{if} $\mathrm{NoGuidance}({\mathcal{O}_i^{\le t}})$ and $\mathrm{AnchorMemoryGatePasses}(c_i,q_i,{\mathcal{O}_i^{\le t}})$ \textbf{then}
\item \hspace*{1.2em}$\Delta_i^{mem} \gets \mathrm{AnchorMemoryCorrection}(c_i,q_i,{\mathcal{O}_i^{\le t}})$
\item \textbf{end if}
\item $a_i \gets b_i\exp(\Delta_i^{mem})$
\item $(d_i^{cur},\sigma_i^{cur}) \gets \mathrm{CurrentEvidence}(E_i,a_i)$
\item $(d_i^{tmp},\sigma_i^{tmp},{\mathcal{M}_i}) \gets \mathrm{TemporalMemory}(E_i,a_i,{\mathcal{O}_i^{\le t}})$
\item \textbf{if} $\mathrm{ExplicitGuidanceEligible}({\mathcal{O}_i^{\le t}},b_i)$ \textbf{then}
\item \hspace*{1.2em}$(d_i^{guid},\sigma_i^{guid}) \gets \mathrm{GuidanceExpert}({\mathcal{O}_i^{\le t}},a_i)$
\item \textbf{else}
\item \hspace*{1.2em}$(d_i^{guid},\sigma_i^{guid}) \gets (0,0)$
\item \textbf{end if}
\item \textbf{for all} ${k} \in {\mathcal{K}}$ \textbf{do}
\item \hspace*{1.2em}${\tilde{\sigma}_i^k} \gets \mathrm{ReliabilityScale}({\sigma_i^k},{k},{\mathcal{O}_i^{\le t}})$
\item \textbf{end for}
\item ${\Omega_i} \gets {\sum_{k\in\mathcal{K}}\tilde{\sigma}_i^k}$
\item \textbf{if} ${\Omega_i} = 0$ \textbf{then}
\item \hspace*{1.2em}${w_i^k} \gets 0$ for each ${k\in\mathcal{K}}$
\item \hspace*{1.2em}${\hat{y}_i^{\mathrm{pre}}} \gets a_i$
\item \hspace*{1.2em}$\hat{y}_i \gets a_i$
\item \textbf{else}
\item \hspace*{1.2em}${w_i^k} \gets {\tilde{\sigma}_i^k} / {\Omega_i}$ for each ${k\in\mathcal{K}}$
\item \hspace*{1.2em}${\hat{y}_i^{\mathrm{pre}}} \gets a_i \exp\left({\sum_{k\in\mathcal{K}} w_i^k d_i^k}\right)$
\item \hspace*{1.2em}$\hat{y}_i \gets \mathrm{GuidanceGuardrail}({\hat{y}_i^{\mathrm{pre}}},a_i,{\mathcal{O}_i^{\le t}})$
\item \textbf{end if}
\item $T_i \gets \mathrm{Trace}(b_i,a_i,\Delta_i^{mem},E_i,{\mathcal{M}_i,\{d_i^k,\sigma_i^k,w_i^k\}_{k\in\mathcal{K}}})$
\item Append the realized row to histories only after $y_i$ becomes observable
\item \textbf{return} $\hat{y}_i,T_i$
\end{list}
\vspace{0.4ex}
\hrule
\end{minipage}
\par\medskip

\subsection{Evidence-Card Interface}
\label{app:evidence-interface}

\begin{tcolorbox}[
breakable,
title={Evidence-Card Interface and Extraction Contract},
colback=gray!3,
colframe=black!45,
boxrule=0.4pt,
arc=1mm,
left=1mm,
right=1mm,
top=1mm,
bottom=1mm,
]
\footnotesize
\noindent\textbf{Input.} Target company, observed and target quarters, allowed company segments, allowed revenue-mechanism relations, and forecast-time document text.
\par
\noindent\textbf{Typed fields.} Source span, affected horizon, target-company segment, relation family, polarity, strength, persistence hint, heuristic confidence, and evidence role.
\par
\noindent\textbf{Revenue filter.} Retain company-attributable demand, supply, pricing, inventory, macro, product-transition, and related revenue mechanisms; reject generic optimism and cost-only commentary.
\par
\noindent\textbf{Source and temporal guard.} Preserve a verbatim span from the supplied document and exclude future documents, realized target actuals, and evidence unavailable by the forecast cutoff.
\par
\noindent\textbf{Validation.} Enforce typed fields, predefined segment and relation labels, source-span availability, target-company consistency, and temporal eligibility.
\par
\noindent\textbf{Output/use.} Return typed JSON cards; validated cards feed Current Evidence and temporal retrieval, while invalid cards abstain from forecasting.
\end{tcolorbox}

\noindent The records are normalized into realized, forward-looking, and delta evidence representations. Company profiles, segment schemas, and guidance schemas declaratively define revenue-relevant drivers and relation families without introducing separate forecasting branches: prompt structure, validation, expert composition, thresholds, and evaluation logic remain shared across companies.
\par\medskip

\subsection{Anchor and Expert Details}
\label{app:anchor-expert-details}

Table~\ref{tab:anchor-family} summarizes the fixed Statistical Anchor candidate family. The family includes classical time-series methods such as autoregressive integrated moving average (ARIMA) and error-trend-seasonal (ETS) models \citep{hyndman2008automatic,HYNDMAN2002439}, a seasonal autoregressive integrated moving-average model with exogenous regressors (SARIMAX) implemented via statsmodels \citep{seabold2010}, and revenue-history plus guidance regressors such as ridge, elastic net, random forest, XGBoost, and LightGBM \citep{hoerl1970ridge,zou2005regularization,breiman2001random,10.1145/2939672.2939785,10.5555/3294996.3295074}. Trainable candidates are fit only on prior rows, and guidance-aware candidates may use management guidance observable at the forecast time. The final matched anchor is selected row by row from this fixed family using prior realized errors under the policy in Table~\ref{tab:anchor-selection-policy}. Missing non-finite anchors are skipped under the evaluation protocol. Before four prior realized rows are available for online candidate scoring, the method uses the first finite positive forecast in this warm-up order: guidance midpoint, same-fiscal-quarter seasonal naive, last-quarter naive, moving average, then historical mean. When direct-guidance candidates lack valid current guidance or sufficient prior observations for fitting, they fall back to the previous realized revenue, preserving complete candidate coverage.

\begin{table*}[t]
\centering
\small
\begin{tabular}{ll}
\toprule
Candidate family & Examples \\
\midrule
Recent-history rules & last-quarter naive; seasonal naive; moving average; drift \\
Classical time-series & ARIMA; ETS \\
Direct guidance anchors & Guidance midpoint; guidance-history blend; guidance affine calibration \\
Revenue-history plus guidance regressors & Linear; Ridge; Elastic Net; Random Forest; XGBoost; LightGBM \\
Time-series with guidance covariate & SARIMAX with guidance \\
\bottomrule
\end{tabular}
\caption{Fixed candidate family for the matched Statistical Anchor.}
\label{tab:anchor-family}
\end{table*}

\begin{table}[t]
\centering
\small
\begin{tabular}{@{}p{0.32\columnwidth}p{0.60\columnwidth}@{}}
\toprule
Property & Value \\
\midrule
Score metric & sMAPE \\
Lookback window for scoring & most recent 24 prior realized rows for the same company \\
Minimum prior history & 4 prior realized rows before history-based selection is used \\
\bottomrule
\end{tabular}
\caption{Statistical Anchor selection policy.}
\label{tab:anchor-selection-policy}
\end{table}

Guidance-gated anchor memory uses prior no-guidance Statistical Anchor errors only. Its shared signed-strength gate uses a threshold of $0.02$ and requires at least six prior no-guidance rows. When the gate passes, the memory branch computes the guarded log-space correction $\Delta_i^{mem}$; otherwise $\Delta_i^{mem}=0$. Anchor memory changes only no-guidance rows; guidance-bearing rows keep $\Delta_i^{mem}=0$ and therefore preserve the selected Statistical Anchor.

For temporal memory, a historical instance $j$ contributes its realized base-anchor residual
\begin{equation}
{\epsilon_j^{hist} = \log y_j - \log a_j.}
\end{equation}
A historical case is scored by typed cross-card attention, recency, same-fiscal-quarter compatibility, and broader context alignment:
\begin{equation}
\begin{aligned}
S(i,j) ={}& A(F_i,F_j)R(i,j) + Q(i,j) \\
&+ \lambda_c A(C_i,C_j)R(i,j),
\end{aligned}
\end{equation}
where $F_i$ and $C_i$ are the forward and broader-context card sets for quarter $i$, $R$ is exponential recency decay, $Q$ is the same-fiscal-quarter bonus, and $A$ is typed card-set matching. At the quarter level, the expert applies a softmax with temperature $T_q=0.35$ over the top $K_q=3$ eligible historical quarters and averages their realized base-anchor residuals.

Typed memory uses transparent card matching based on explicit type features. Given a query card $x$ and a memory card $z$, we define
\begin{equation}
{
\begin{aligned}
\phi(x,z) ={}& 0.30\,\mathbf{1}_{seg}(x,z) + 0.25\,\mathbf{1}_{rel}(x,z) \\
&+ 0.15\,P(x,z) + 0.10\,H(x,z) \\
&+ 0.10\,U(x,z) + 0.10\,J(x,z),
\end{aligned}
}
\end{equation}
Here $\mathbf{1}_{seg}$ and $\mathbf{1}_{rel}$ indicate exact normalized segment and relation-family matches. $P(x,z)$ equals $1$ for matching nonzero polarities, $0$ for opposing polarities, and $0.5$ if either polarity is mixed or indeterminate; $H(x,z)$ is the binary persistence-hint match; $U(x,z)=\max\{0,1-|\nu(s_x)-\nu(s_z)|\}$ with $\nu(\mathrm{low})=0.33$, $\nu(\mathrm{medium})=0.66$, $\nu(\mathrm{high})=1$, and $\nu(\mathrm{other})=0.5$; and $J$ is token Jaccard overlap over the evidence span. For card sets $X$ and $Z$, we compute
\begin{equation}
{
\begin{aligned}
A(X,Z) ={}& \frac{\sum_{x\in X}\omega_x \sum_{z\in T_{K_a}(x,Z)} \beta_{xz}\phi(x,z)}{\sum_{x\in X}\omega_x}, \\
\beta_{xz} ={}& \frac{\exp(\phi(x,z)/T_a)}{\sum_{z'\in T_{K_a}(x,Z)}\exp(\phi(x,z')/T_a)}.
\end{aligned}
}
\end{equation}
$T_{K_a}(x,Z)$ is the set of up to $K_a$ cards in $Z$ with the largest $\phi(x,z)$, and $\omega_x$ is the nonnegative query-card salience induced by polarity, strength, and extraction confidence. We define $A(X,Z)=0$ when either set is empty or total query salience is zero. The coefficients in $\phi$ are shared manual choices that prioritize segment and relation-family agreement before polarity, persistence, strength, and lexical overlap. The within-quarter matcher uses $K_a=2$, $T_a=0.20$, and $\lambda_c=0.35$, distinct from the quarter-retrieval settings $K_q$ and $T_q$. The Temporal-Memory alignment guard applies only to weak derived numeric guidance and multiplies both the temporal residual and its support by the smaller of the mean- and top-match direction-alignment scores.

The current-evidence channel proposes $d_i^{cur}$ with raw reliability $\sigma_i^{cur}$ and clipped reliability $\bar{\sigma}_i^{cur}=\mathrm{clip}(\sigma_i^{cur},0,1)$. We compute a prior-history scale $\kappa_i^{cur}$ and set $\tilde{\sigma}_i^{cur}=\kappa_i^{cur}\bar{\sigma}_i^{cur}$ before Eq.~\ref{eq:came-aggregation}. The history bucket is defined by guidance availability, current/temporal sign agreement, conflict level, and explicit-guidance activity. Prior rows are selected in the fixed order of the exact four-way bucket, the matching guidance-availability bucket, and then all usable prior rows; if fewer than four rows remain, the scale defaults to one. Let $\bar{e}_i$ be the prior mean marginal sMAPE effect and $p_i^{win}$ the prior win rate after this fallback order. With minimum history $4$, $\tau_{cur}=0.01$, $\kappa_{min}=0.25$, and $\kappa_{max}=1.35$,
\begin{equation}
{
\begin{aligned}
\zeta_i^{cur} ={}& \mathrm{clip}\Bigl(0.65\tanh(\bar{e}_i/\tau_{cur}) \\
&+ 0.35(2p_i^{win}-1), -1, 1\Bigr).
\end{aligned}
}
\end{equation}
The scale is $\kappa_i^{cur}=1+\zeta_i^{cur}(\kappa_{max}-1)$ for $\zeta_i^{cur}\ge0$ and $\kappa_i^{cur}=1+\zeta_i^{cur}(1-\kappa_{min})$ otherwise. Explicit numeric-guidance rows apply an additional $0.65$ multiplier to $\kappa_i^{cur}$ to reduce overlap with the guidance correction; if the resulting scale is below one, the current-evidence delta is scaled by the same factor.

The $\tanh$ transform maps prior marginal effect monotonically to a reliability adjustment, while win rate supplies a separate stability signal.

For explicit numeric guidance, the guidance residual channel proposes
\begin{equation}
{
\begin{aligned}
d_i^{guid} &= \log \frac{g_i^{mid}}{a_i}, \\
\sigma_i^{guid} &= \mathrm{clip}\Bigl(
\gamma_g\,\ell_i \exp(-e_i^{guid}/\tau)\eta_i^{hist}, 0, 1\Bigr), \\
\eta_i^{hist} &= \min\left(\frac{n_i}{m_g}, 1\right).
\end{aligned}
}
\end{equation}
Here $g_i^{mid}$ is the current explicit guidance midpoint, $e_i^{guid}$ is the prior-history absolute log anchor error (same-guidance history when available, then recent or overall history), and $n_i$ is the number of available prior anchor-error rows. Let $q_i^{guid}$ denote the row-level heuristic guidance-quality score. The guidance-quality factor $L_i\in[0,1]$ is the clipped product of binary eligible explicit-numeric availability, $\mathrm{clip}(q_i^{guid}/20,0,1)$, and $\max\{0,1-B_i/0.15\}$, where $B_i=|g_i^{high}-g_i^{low}|/|g_i^{mid}|$ is relative guidance-band width. We use $m_g=1$, $\tau=0.20$, $\gamma_g=1.25$, and $\ell_i=\mathrm{clip}(0.35+0.65L_i,0,1)$.

The Guidance Guardrail applies
\begin{equation}
{\hat{y}_i = a_i + \alpha(\chi_i^{guid})(\hat{y}_i^{\mathrm{pre}} - a_i).}
\end{equation}
Here $\chi_i^{guid}$ denotes the forecast-time guidance category, with $\alpha=1.0$ for Explicit Numeric Guidance and No Guidance, $\alpha=0.0$ for weak derived numeric guidance (a subset of Non-Explicit Guidance), and $\alpha=0.5$ for the remaining Non-Explicit Guidance categories of forward commentary or qualitative-only guidance. All forecasting-active constants are shared across firms.

\paragraph{Parameter Selection and Sensitivity.}
The settings $M_{\mathrm{AM}}=6$, $\tau_{\mathrm{AM}}=0.02$, $K_q=3$, and $T_q=0.35$ were selected during development on AAPL, NVDA, and AVGO through FY2023 Q4. The historical development record does not support a more granular rationale for each exact value; Table~\ref{tab:memory-sensitivity} therefore reports one-factor alternatives. The card-matcher, reliability, and guardrail coefficients are shared heuristic settings. All settings were fixed before temporal and company-held-out evaluation and were applied without per-company tuning. The study does not claim exhaustive hyperparameter tuning.

\section{Baseline and Prompt Protocols}
\label{app:prompt-comparator-details}

\subsection{Model and Input Protocols}

Closed-model calls use the official OpenAI API snapshots \texttt{gpt-4o-mini-2024-07-18} for taxonomy discovery \citep{openai2024gpt4omini} and \texttt{gpt-4.1-mini-2025-04-14} for evidence relation/event extraction and the main-table direct LLM baselines \citep{openai2025gpt41}. Evidence event extraction and the direct baselines used temperature 0.0; relation extraction omitted temperature and therefore used the provider default. The provider reports a June 2024 knowledge cutoff for \texttt{gpt-4.1-mini-2025-04-14}. Because the rolling benchmark contains outcomes both before and after this cutoff, potential pretrained exposure cannot be ruled out for the full panel, and cutoff timing alone does not establish whether any individual prediction relied on memorized company information.

For History-only and History + Guidance, the target-quarter actual is hidden, and only the specified forecast-time inputs are provided. We replace only the company-name token with \texttt{Entity X} to reduce direct company-identity cues. However, exact fiscal periods, ISO 4217 reporting currency, absolute revenue scale, and historical trajectories remain unchanged and may still permit company re-identification. We therefore treat this masking only as a reduction of explicit identity cues, not as a guarantee against re-identification or reliance on company-specific pretrained knowledge.

History-only receives an expanding contiguous sequence of prior realized quarters beginning with an FY2018 Q1 warm-up and returns strict two-field JSON. History + Guidance additionally receives prior guidance and numeric target-quarter guidance when available: it generates forecasts for 193 rows and reuses the corresponding History-only forecast on 143 rows; the temporal held-out counts are 62 and 34. This yields complete coverage without imputing missing guidance. CAME's Explicit Numeric slice denotes direct numeric management guidance, whereas History + Guidance uses a separate binary target-guidance rule that can also admit derived numeric ranges; the category counts therefore need not coincide.

TimeGPT-1 (API model \texttt{timegpt-1}, accessed 2026-07-10) uses a contiguous prior-quarter suffix for each one-step forecast \citep{garza2024timegpt}. Zero-shot Chronos-Bolt Base (\texttt{amazon/chronos-bolt-base}, 205M parameters) uses all finite prior revenues \citep{ansari2024chronos}. Both cover 336/336 rows without guidance or text. Supplementary open-weight baselines ran on one NVIDIA RTX A5000 24GB GPU using \texttt{mistral-small:24b}, \texttt{llama3.1:8b}, and \texttt{Qwen/Qwen2.5-14B-Instruct-AWQ} \citep{mistralai2025small3,grattafiori2024llama3herd,Yang2024Qwen25TR}.

All 529 required identity-masked direct-LLM calls completed successfully without retries or cache reuse.

\subsection{Exact Direct-Baseline Prompt Templates}

The following boxes reproduce the fixed History-only and History + Guidance prompts. Angle-bracketed placeholders denote row-specific values and CSV blocks; \texttt{Entity X} is literal.

\begin{tcolorbox}[
title={Exact System Prompt},
colback=gray!3,
colframe=black!45,
coltext=black,
boxrule=0.4pt,
arc=1mm,
left=1mm,
right=1mm,
top=1mm,
bottom=1mm,
breakable
]
\scriptsize\ttfamily
\detokenize{You are a financial forecasting assistant.}\par
\detokenize{You will be given Entity X's historical quarterly data and optionally management guidance.}\par
\detokenize{You must forecast the revenue for ONE target quarter.}\par
\medskip
\detokenize{Rules:}\par
\detokenize{- Use only the provided tables; do not assume you know any future realized values.}\par
\detokenize{- Output MUST follow the required JSON schema (no extra keys, no commentary).}\par
\detokenize{- Return a single object with fields fiscal_quarter and pred_revenue.}
\end{tcolorbox}

\begin{tcolorbox}[
title={Exact History-only User Template},
colback=gray!3,
colframe=black!45,
coltext=black,
boxrule=0.4pt,
arc=1mm,
left=1mm,
right=1mm,
top=1mm,
bottom=1mm,
breakable
]
\scriptsize\ttfamily
\detokenize{Below is Entity X's historical quarterly revenue (realized).}\par
\detokenize{The unit of all values is <UNIT_DESC>.}\par
\medskip
\detokenize{Historical data (past quarters):}\par
\detokenize{```text}\par
\detokenize{<HISTORY_CSV>}\par
\detokenize{```}\par
\medskip
\detokenize{Task: Forecast the realized revenue for the target quarter: <TARGET_FISCAL_QUARTER>.}\par
\detokenize{Return only the JSON object.}
\end{tcolorbox}

\begin{tcolorbox}[
title={Exact History + Guidance User Template},
colback=gray!3,
colframe=black!45,
coltext=black,
boxrule=0.4pt,
arc=1mm,
left=1mm,
right=1mm,
top=1mm,
bottom=1mm,
breakable
]
\scriptsize\ttfamily
\detokenize{Below is Entity X's historical quarterly data.}\par
\detokenize{The unit of all revenue and guidance values is <UNIT_DESC>.}\par
\medskip
\detokenize{Historical data (past quarters; realized revenue + guidance):}\par
\detokenize{```text}\par
\detokenize{<HISTORY_WITH_GUIDANCE_CSV>}\par
\detokenize{```}\par
\medskip
\detokenize{For the target quarter <TARGET_FISCAL_QUARTER>, you ONLY see management guidance (revenue is unknown):}\par
\detokenize{```text}\par
\detokenize{<TARGET_GUIDANCE_CSV>}\par
\detokenize{```}\par
\medskip
\detokenize{Task: Using the historical relationship between guidance and realized revenue (bias, under/over-shoot, etc.), forecast the realized revenue for <TARGET_FISCAL_QUARTER>.}\par
\detokenize{Return only the JSON object.}
\end{tcolorbox}

\begin{table*}[t]
\centering
\scriptsize
\resizebox{\textwidth}{!}{%
\begin{tabular}{lrlrrrrr}
\toprule
Company & Rows & Guidance mix & Statistical Anchor & History + Guidance & CAME & $\Delta$ vs anchor & $\Delta$ vs LLM \\
\midrule
AAPL & 28 & 6 / 14 / 8 & 0.071 / 5.99B & 0.079 / 6.77B & 0.057 / 4.97B & +0.014 & +0.022 \\
NVDA & 28 & 28 / 0 / 0 & 0.051 / 0.37B & 0.071 / 0.80B & 0.051 / 0.35B & +0.001 & +0.021 \\
AVGO & 28 & 19 / 0 / 9 & 0.048 / 0.37B & 0.028 / 0.26B & 0.033 / 0.27B & +0.016 & -0.005 \\
AMZN & 28 & 24 / 1 / 3 & 0.062 / 6.28B & 0.046 / 4.93B & 0.043 / 4.13B & +0.019 & +0.003 \\
ASML & 28 & 27 / 0 / 1 & 0.055 / 0.25B & 0.058 / 0.28B & 0.051 / 0.23B & +0.004 & +0.007 \\
GOOGL & 28 & 0 / 10 / 18 & 0.091 / 5.39B & 0.065 / 3.67B & 0.082 / 4.81B & +0.009 & -0.017 \\
INTC & 28 & 17 / 2 / 9 & 0.064 / 1.08B & 0.064 / 1.04B & 0.054 / 0.89B & +0.010 & +0.010 \\
META & 28 & 17 / 5 / 6 & 0.086 / 2.15B & 0.070 / 1.99B & 0.064 / 1.60B & +0.022 & +0.006 \\
MSFT & 28 & 0 / 10 / 18 & 0.044 / 1.81B & 0.041 / 1.91B & 0.035 / 1.37B & +0.009 & +0.006 \\
MU & 28 & 27 / 0 / 1 & 0.050 / 0.31B & 0.055 / 0.36B & 0.041 / 0.25B & +0.010 & +0.015 \\
TSLA & 28 & 0 / 0 / 28 & 0.173 / 2.54B & 0.159 / 2.25B & 0.162 / 2.43B & +0.010 & -0.003 \\
ORCL & 28 & 0 / 23 / 5 & 0.043 / 0.48B & 0.050 / 0.59B & 0.039 / 0.42B & +0.005 & +0.011 \\
\bottomrule
\end{tabular}%
}
\caption{Per-company sMAPE / MAE heterogeneity on the development-inclusive 336-row rolling backtest. Guidance mix lists CAME's Explicit / Non-explicit / No-guidance categories.}
\label{tab:per-company-heterogeneity}
\end{table*}

{
\noindent The substitution and validation contract is:
\begin{itemize}[leftmargin=*,nosep]
\item \texttt{<UNIT\_DESC>} is \texttt{billion USD}, except for ASML, where it is \texttt{billion EUR}. \texttt{<TARGET\_FISCAL\_QUARTER>} is the target label.
\item \texttt{<HISTORY\_CSV>} has columns \texttt{fiscal\_quarter,revenue}. \texttt{<HISTORY\_WITH\_GUIDANCE\_CSV>} has columns \texttt{fiscal\_quarter,}\allowbreak\texttt{revenue,}\allowbreak\texttt{guid\_low,}\allowbreak\texttt{guid\_high,}\allowbreak\texttt{guid\_mid,}\allowbreak\texttt{pct}. Both contain the expanding contiguous prior-quarter history beginning at FY2018 Q1.
\item \texttt{<TARGET\_GUIDANCE\_CSV>} contains one row with \texttt{fiscal\_quarter}, \texttt{guid\_low}, \texttt{guid\_high}, \texttt{guid\_mid}, and \texttt{pct}. Monetary values use the displayed billion-unit scale; \texttt{pct} remains on a 0--100 scale, and missing historical guidance is empty.
\item The returned JSON contains only \texttt{fiscal\_quarter} and \texttt{pred\_revenue}; parsing requires the target quarter and a finite positive prediction.
\item History + Guidance is called only when target-quarter low, midpoint, and high guidance are finite and ordered. If all three and \texttt{pct} are absent, the corresponding History-only result is reused; partial, nonfinite, incorrectly ordered, or percentage-only guidance fails validation.
\end{itemize}
}

\subsection{Text Comparator Setup}

Transcript-only uses the observed-quarter earnings call; CARAG-style retrieval adds retrieved same-company historical transcript cases; and Wide-context revenue adds historical revenue, guidance, transcript features, and segment summaries. Filing-only uses forecast-time filing text and MD\&A-only its MD\&A section. History + Transcript combines prior realized revenue with the full observed-quarter call, without a separate guidance table, to predict target-quarter total revenue.

\section{Additional Results and Diagnostics}

\subsection{Per-Company Results and Forecast Trajectories}
\label{app:rq1-additional-results}

Table~\ref{tab:per-company-heterogeneity} gives the company-level readout behind the aggregate result. CAME improves over the matched Statistical Anchor for every company by sMAPE. Against History + Guidance, CAME has lower company sMAPE for nine of 12 firms; AVGO, GOOGL, and TSLA favor the direct comparator. Positive deltas are comparator sMAPE minus CAME sMAPE, so positive values favor CAME.

\begin{figure*}[t]
\centering
\includegraphics[width=0.82\textwidth]{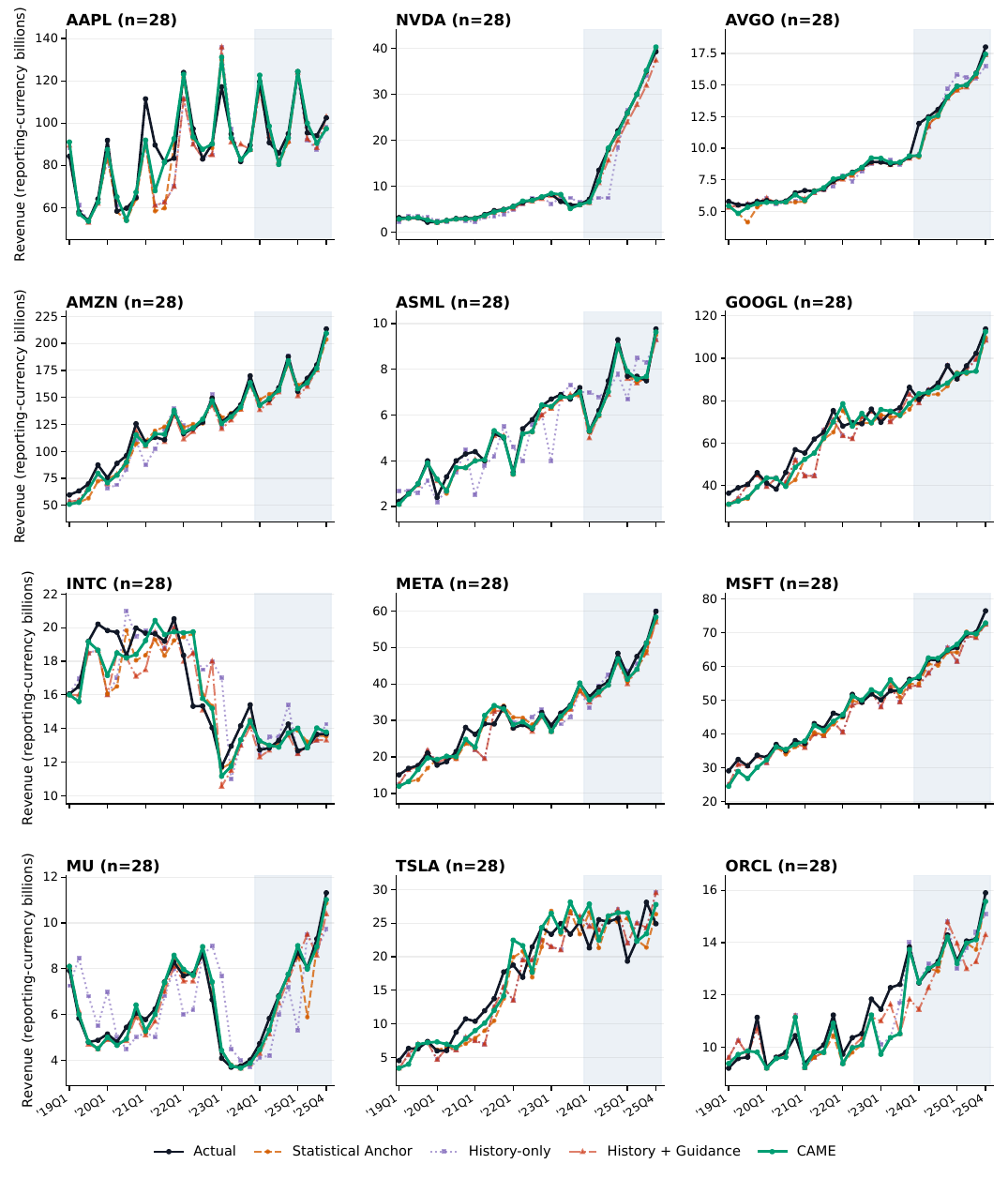}
\caption{Comparator trajectories for all 12 firms; shading marks the 96-row temporal held-out window and axes use company-local reporting-currency billions.}
\label{fig:app-rq1-comparator-trajectories}
\end{figure*}

Figure~\ref{fig:app-rq1-comparator-trajectories} provides the corresponding quarter-level trajectories for the Statistical Anchor, both direct LLM variants, CAME, and realized revenue.

\onecolumn
\twocolumn
\section{Extended Experiments}

\subsection{Sensitivity, Stress, and Validation Analyses}
\label{app:sensitivity-validation}

Table~\ref{tab:memory-sensitivity} changes one shared setting from $M_{\mathrm{AM}}=6$, $\tau_{\mathrm{AM}}=0.02$, $K_q=3$, and $T_q=0.35$; the paired $M_{\mathrm{AM}}$ and $\tau_{\mathrm{AM}}$ alternatives produce identical predictions. Its Anchor--CAME column is Statistical Anchor minus CAME, so positive values favor CAME. Table~\ref{tab:upstream-ablation} compares a strict extraction rule requiring a distinct revenue driver with a broader rule that also admits forward revenue outlooks without one. Each rule has two stochastic replicates with common sources, models, and subsequent settings; all variants beat the matched anchor. Broader-minus-strict differences indicate directional sensitivity rather than a population confidence interval. Table~\ref{tab:extraction-stress} averages three outcome-blind seeds at each synthetic rate. Its sMAPE deltas are stressed minus unstressed CAME, while guidance attenuation measures reduced absolute prediction movement on eligible changed rows rather than a guard-disabled effect or real-world prevention rate.

\begin{table*}[t]
\centering
\scriptsize
{
\begin{tabular}{lrrrr}
\toprule
Setting & Development-Inclusive (336) sMAPE & Temporal Held-Out (96) sMAPE & No-guidance sMAPE & Anchor--CAME (336) \\
\midrule
CAME & 0.059253 & 0.037810 & 0.085681 & +0.010697 \\
$M_{\mathrm{AM}}=4$ or $8$ & 0.059253 & 0.037810 & 0.085681 & +0.010697 \\
$\tau_{\mathrm{AM}}=0.01$ or $0.04$ & 0.059253 & 0.037810 & 0.085681 & +0.010697 \\
$K_q=1$ & 0.059320 & 0.038634 & 0.085910 & +0.010630 \\
$K_q=5$ & 0.059557 & 0.038423 & 0.085607 & +0.010393 \\
$T_q=0.20$ & 0.059180 & 0.037877 & 0.085624 & +0.010770 \\
$T_q=0.50$ & 0.059305 & 0.037799 & 0.085714 & +0.010646 \\
\bottomrule
\end{tabular}
}
\caption{\textbf{One-factor memory and retrieval sensitivity.}}
\label{tab:memory-sensitivity}
\end{table*}

\begin{table*}[t]
\centering
\scriptsize
{
\begin{tabular}{lrrrrrr}
\toprule
Replicate & Strict Full & Broader Full & Full Difference & Strict Held-Out & Broader Held-Out & Held-Out Difference \\
\midrule
1 & 0.059686 & 0.059129 & -0.000557 & 0.038409 & 0.038580 & +0.000171 \\
2 & 0.060097 & 0.059607 & -0.000490 & 0.038368 & 0.039171 & +0.000803 \\
Mean & 0.059892 & 0.059368 & -0.000524 & 0.038389 & 0.038876 & +0.000487 \\
\bottomrule
\end{tabular}
}
\caption{\textbf{Evidence-extraction sensitivity.}}
\label{tab:upstream-ablation}
\end{table*}

\begin{table*}[t]
\centering
\scriptsize
{
\begin{tabular}{llrrrr}
\toprule
Stress & Rate & Full $\Delta$ & Held-out (96) $\Delta$ & Held-out (96) Anchor--CAME & Guidance attenuation \\
\midrule
Polarity flip & 5\% & +0.000105 & +0.000341 & +0.009019 & 77.1\% \\
Polarity flip & 10\% & +0.000202 & +0.000500 & +0.008860 & 71.7\% \\
Forward-card omission & 5\% & -0.000048 & +0.000162 & +0.009198 & 69.5\% \\
Forward-card omission & 10\% & +0.000062 & +0.000401 & +0.008959 & 67.2\% \\
False-positive horizon & 5\% & -0.000007 & +0.000351 & +0.009009 & 65.1\% \\
False-positive horizon & 10\% & +0.000096 & +0.000209 & +0.009151 & 66.1\% \\
\bottomrule
\end{tabular}
}
\caption{\textbf{Controlled evidence-card error stress.}}
\label{tab:extraction-stress}
\end{table*}

These interventions show partial rather than universal protection. Aggregate development-inclusive and temporal held-out margins over the matched anchor remain positive, but one 10\% seed in each stress family yields a negative anchor margin on the non-explicit-guidance slice (Anchor--CAME $=-0.002263/-0.000588/-0.000064$ for polarity flip, omission, and false-positive horizon). The Guidance Guardrail attenuates all measured harmful false-positive-horizon occurrences and produces zero stress-induced prediction movement in 31/54 and 37/62 pooled synthetic occurrences at 5\% and 10\%, respectively, while the Temporal-Memory alignment guard amplifies one polarity setting. These synthetic counts are neither population error rates nor real-world prevention probabilities.

\subsection{Human Annotation Protocol and Instructions}
\label{app:human-annotation-protocol}

The fixed 120-card sample was balanced across 12 companies and stratified by period. Each record showed authoritative English source context, the extracted factor, and company/date/observed-quarter/target-quarter metadata; line-aligned Traditional Chinese was a reading aid. System labels, outcomes, predictions, errors, diagnostics, and other responses were hidden. No demographic, sensitive, or other personal information was collected for analysis.

Two primary evaluators independently labeled all 120 cards using the displayed fiscal metadata and authoritative English source. They did not discuss cards or view predictions, actuals, errors, diagnostics, system labels, or each other's responses.

\begin{tcolorbox}[
title={Primary-Evaluator Question Prompts},
colback=gray!3,
colframe=black!45,
boxrule=0.4pt,
arc=1mm,
left=1mm,
right=1mm,
top=1mm,
bottom=1mm,
breakable
]
\small
\begin{enumerate}[leftmargin=*,nosep]
\item \textbf{Source grounding.} ``Does the displayed factor accurately summarize what the source says? Judge summary faithfulness, not revenue relevance.'' Labels: Fully supported, Partially supported, Unsupported, and Insufficient context.
\item \textbf{Revenue validity.} ``Does the source support a plausible path by which this factor affects the target company's revenue?'' Labels: Yes, No, and Unclear. A Yes required a target-company revenue path through mechanisms such as demand, shipments, price, usage, or renewals; cost, margin, capital expenditure, earnings-per-share, general-market, or other-company information alone was No.
\item \textbf{Revenue direction.} ``If the source supports a target-company revenue impact, what direction does it support? Do not answer from cost, margin, profit, or another entity.'' Labels: Positive, Negative, Mixed, No direction, and Unclear.
\item \textbf{Next-quarter support.} ``Does the evidence support the next fiscal quarter after the displayed observed quarter?'' Labels: Yes, Partially, No, and Unclear. Yes required explicit or metadata-aligned next-quarter timing; Partially covered the next quarter plus other periods or imprecise timing; No denoted current-quarter or historical evidence.
\item \textbf{Correction rationale.} ``Briefly explain any unsupported, no, no-direction, partial, unclear, or insufficient-context judgment.'' This text field was optional.
\end{enumerate}
\end{tcolorbox}

A third evaluator adjudicated only disputed dimensions from the English source and context. Primary order was anonymized and randomized; outcomes, predictions, errors, diagnostics, system labels, and non-adjudication records remained hidden. All evaluators were internal (two authors and one laboratory colleague). Participation was voluntary and uncompensated; no payment or other material compensation was provided. Before annotation, all evaluators were informed that their judgments would be used for research and reported in this paper, and they agreed to participate on that basis. The results assess sampled card semantics, not causal correctness or trace utility.

Strict derived card validity requires source support (full or partial), target-company revenue validity, an evaluable revenue direction, and strict target-quarter support. Partial-inclusive validity applies the same conditions but also accepts partial target-quarter support.

Table~\ref{tab:human-card-audit} summarizes primary-evaluator agreement and the adjudicated results.

\begin{table*}[t]
\centering
\footnotesize
\setlength{\tabcolsep}{3pt}
\renewcommand{\arraystretch}{1.05}
{
\begin{tabular}{@{}>{\raggedright\arraybackslash}p{0.16\textwidth}>{\raggedright\arraybackslash}p{0.20\textwidth}>{\raggedright\arraybackslash}p{0.10\textwidth}>{\raggedright\arraybackslash}p{0.47\textwidth}@{}}
\toprule
Dimension & Primary-evaluator exact agreement & Cohen's $\kappa$ & Adjudicated result / system agreement \\
\midrule
Source grounding & 88/120 (73.3\%) & 0.282 & 102 full; 18 partial (overstatement, omitted qualification, or scope/subject mismatch) \\
Target-company revenue validity & 81/120 (67.5\%) & 0.269 & 96/120 valid \\
Revenue direction & 64/120 (53.3\%) & 0.356 & System polarity vs. consensus: 86/110 exact; 10 strict reversals; 14 mixed/one-sided mismatches; 10 non-evaluable (2 no-direction; 8 unclear) \\
Target-quarter support & 74/120 (61.7\%) & 0.443 & 55 strict; 41 partial; 24 unsupported \\
Derived card validity & -- & -- & 49 strict; 84 partial-inclusive \\
\bottomrule
\end{tabular}
}
\caption{\textbf{Outcome-blind internal evaluation of 120 evidence cards.}}
\label{tab:human-card-audit}
\end{table*}

\subsection{Failure-Pattern Diagnostic}

Table~\ref{tab:failure-patterns} applies a deterministic ordered rule to the 12 highest-harm cases selected for each of four components (48 component-case selections; 46 unique forecasts). Categories are exclusive, mechanisms may overlap, and the counts are not prevalence estimates.

\subsection{Incremental Processing Cost}

Table~\ref{tab:incremental-cost} reports one 12-company quarterly update from FY2025 Q3 to FY2025 Q4. The 7.00-minute total uses two concurrent workers and excludes the separately timed replay. Usage cost is computed from returned API usage and provider rates checked on 2026-07-12. The profile excludes source acquisition, schema setup, human quality assurance, comparators, historical backfill, and unobserved client-library retries, so it is not an end-to-end deployment estimate.

\begin{table*}[t]
\centering
\begin{minipage}[t]{0.47\textwidth}
\centering
\scriptsize
{
\begin{tabular}{@{}p{0.72\linewidth}r@{}}
\toprule
Failure category & Count \\
\midrule
Guidance-residual misalignment & 18 \\
Anchor-Memory + Temporal over-correction & 8 \\
Temporal duplicate/amplification & 8 \\
Anchor-Memory over-correction & 4 \\
Current-Evidence residual mismatch & 4 \\
Temporal direction conflict & 3 \\
Temporal over-correction & 3 \\
\bottomrule
\end{tabular}
}
\captionof{table}{\textbf{Failure-pattern diagnostic.}}
\label{tab:failure-patterns}
\end{minipage}
\hfill
\begin{minipage}[t]{0.47\textwidth}
\centering
\scriptsize
{
\begin{tabular}{@{}p{0.68\linewidth}r@{}}
\toprule
Quantity & Measured value \\
\midrule
Successful API calls & 80 \\
Prompt / completion tokens & 530,311 / 46,637 \\
Usage-priced API cost & \$0.244681 \\
Evidence-extraction time & 404.37 s \\
Evidence-card construction and CAME scoring time & 15.78 s \\
Total incremental processing time & 420.15 s (7.00 min) \\
Separate post-extraction replay & 15.81 s \\
\bottomrule
\end{tabular}
}
\captionof{table}{\textbf{Measured quarterly update cost for 12 firms.}}
\label{tab:incremental-cost}
\end{minipage}
\end{table*}

\subsection{Bootstrap Uncertainty Analysis}
\label{app:rq4-bootstrap-details}

\begin{table*}[t]
\centering
\resizebox{\textwidth}{!}{%
\begin{tabular}{lrrrrrl}
\toprule
Surface & $N$ & Companies & Statistical Anchor sMAPE & CAME sMAPE & $\Delta$ sMAPE & Macro 95\% CI \\
\midrule
Development-inclusive rolling backtest & 336 & 12 & 0.069950 & 0.059253 & +0.010697 & [+0.007463, +0.014067] \\
Temporal held-out & 96 & 12 & 0.047171 & 0.037810 & +0.009360 & [+0.004235, +0.016219] \\
Company-held-out temporal subset & 72 & 9 & 0.046687 & 0.036616 & +0.010071 & [+0.003458, +0.018824] \\
Explicit Numeric Guidance & 165 & 8 & 0.041196 & 0.036894 & +0.004302 & [-0.001523, +0.009701] \\
Non-Explicit Guidance & 65 & 7 & 0.099437 & 0.098112 & +0.001324 & [-0.000721, +0.003305] \\
No Guidance & 106 & 11 & 0.144171 & 0.085681 & +0.058490 & [+0.023849, +0.111277] \\
\bottomrule
\end{tabular}%
}
\caption{Paired-bootstrap uncertainty against the Statistical Anchor.}
\label{tab:bootstrap}
\end{table*}

Table~\ref{tab:bootstrap} provides the paired-bootstrap results summarized in the main text. Delta is Statistical Anchor minus CAME, so positive values favor CAME. The development-inclusive rolling backtest and no-guidance slice have strictly positive intervals, while the two guidance-bearing slices have positive sMAPE point estimates but intervals that touch or cross zero. All intervals resample company-level macro-sMAPE deltas with replacement for 10,000 draws. The 336-row and 96-row surfaces are evaluated separately; the temporal held-out and company-held-out temporal analyses resample 12 and 9 company clusters, respectively. Statistical Anchor contrasts use seed 43; the four direct-LLM contrasts use seeds 20260827--20260830 in the order reported. Against the direct LLM baselines, development-inclusive intervals are +0.038517 $[+0.010691,+0.071672]$ for History-only and +0.006316 $[+0.000264,+0.011813]$ for History + Guidance; temporal held-out intervals are +0.036096 $[+0.003522,+0.075153]$ and +0.014261 $[-0.002088,+0.030635]$, respectively. Thus only temporal held-out History + Guidance crosses zero.

\section{Traceability Case Summary}
\label{app:rq5-traceability}
Figures~\ref{fig:app-aapl-trace-audit} and~\ref{fig:app-memory-boundary-cases} contrast an AAPL lower-error case with a TSLA high-anchor boundary case. Both use author-written normalized source-linked summaries rather than verbatim source text. Retrospective target-quarter cards are shown only for post-hoc interpretation and never enter forecasting.

\begin{figure*}[t]
\centering
\begin{tcolorbox}[colback=blue!2,colframe=blue!55,coltitle=black,colbacktitle=blue!7,title=\textbf{AAPL FY2024 Q4 Trace Card: Anchor $\rightarrow$ Evidence/Memory $\rightarrow$ Guarded Output},fonttitle=\bfseries,boxrule=0.5pt,arc=2pt,left=4pt,right=4pt,top=4pt,bottom=4pt]
\scriptsize
\setlength{\tabcolsep}{3pt}
\renewcommand{\arraystretch}{1.13}

\textbf{Panel A. Forecast path and boundary.}\\[2pt]
\begin{tabular}{p{0.18\linewidth}p{0.27\linewidth}p{0.47\linewidth}}
\toprule
Step & Value & Interpretation \\
\midrule
Forecast boundary & FY2024 Q3 $\rightarrow$ FY2024 Q4 & Forward-looking commentary; no explicit total-revenue guidance; inputs are forecast-time only. \\
Statistical Anchor & 90.93B; base anchor unchanged & Base anchor equals the Statistical Anchor; anchor memory is inactive on this guidance-bearing row. \\
Evidence-memory update & 95.11B pre-guardrail & iPhone/Apple Intelligence and Services cards plus temporal analogs support a positive correction, moderated by FX headwind. \\
Guarded CAME output & 93.02B CAME; 94.93B actual & Guardrail keeps half the positive correction for forward commentary; actual is evaluation-only. \\
\bottomrule
\end{tabular}

\vspace{6pt}
\textbf{Panel B. Forecast-time \textit{forward-looking/guidance} evidence cards.} These cards explain the correction. They are presented as author-written normalized summaries of source-linked forecast-time management commentary; verbatim transcript text is not reproduced.\\[2pt]
\begin{tabular}{p{0.05\linewidth}p{0.15\linewidth}p{0.13\linewidth}p{0.17\linewidth}p{0.42\linewidth}}
\toprule
Card & Segment / factor & Signal & Role in trace & Author-written source-linked summary \\
\midrule
C1 & iPhone / Apple Intelligence & positive demand & supports positive correction & Apple Intelligence was tied to advanced iPhone models and presented as a new capability set. \\
C2 & Services / services growth & positive revenue & supports positive correction & Management expected Services revenue to continue growing at a double-digit rate similar to the first three fiscal quarters. \\
C3 & Company-level / foreign exchange & negative headwind & conflict signal; moderates confidence & Management expected foreign exchange to remain a revenue headwind of about 1.5 percentage points year over year. \\
\bottomrule
\end{tabular}

\vspace{6pt}
\textbf{Panel C. Forecast-time-eligible temporal memory used by the trace.} The temporal expert retrieves AAPL analog quarters whose realized anchor-relative residuals were positive. Residual cues come from prior realized outcomes, not target-quarter evidence.\\[2pt]
\begin{tabular}{p{0.16\linewidth}p{0.11\linewidth}p{0.18\linewidth}p{0.47\linewidth}}
\toprule
Retrieved quarter & Weight & Residual cue & Why it is shown \\
\midrule
FY2023 Q4 & 0.4049 & +0.0235 log & Primary temporal analog; strong typed/semantic alignment with current C3 on FX-headwind evidence (match 0.983) and a positive realized residual. \\
FY2024 Q3 & 0.3953 & +0.0645 log & Recent same-company temporal proxy; its positive realized residual indicates prior anchor underprediction in a nearby context. \\
FY2022 Q4 & 0.1999 & +0.0218 log & Older same-company analog providing weaker but directionally consistent positive residual evidence. \\
\bottomrule
\end{tabular}

\vspace{6pt}
\textbf{Panel D. Retrospective card-alignment check for target FY2024 Q4.} These post-report target-quarter cards are used only for interpretation; they are not part of the forecast-time evidence set and never change predictions, weights, gates, comparator rows, or metrics.\\[2pt]
\begin{tabular}{p{0.12\linewidth}p{0.22\linewidth}p{0.20\linewidth}p{0.36\linewidth}}
\toprule
Alignment marker & Realized target-quarter driver & Matched forecast-time card & Interpretation note \\
\midrule
\textcolor{green!45!black}{\textbf{exact match}} & iPhone / Apple Intelligence & C1; exact driver match & Target-quarter report says Apple Intelligence features became available for iPhone, iPad, and Mac. \\
\textcolor{green!45!black}{\textbf{segment-factor match}} & Services / Apple Pay usage & C2; segment-factor alignment & Target-quarter report cites Apple Pay adoption as a Services-related realized driver. \\
\textcolor{red!60!black}{\textbf{unmatched driver}} & Mac / M4 chip transition & Absence of forecast-time Mac-related evidence & Notes an unanticipated product-cycle driver absent from forecast-time texts, helping localize one source of residual forecast error. \\
\bottomrule
\end{tabular}
\end{tcolorbox}
\caption{AAPL FY2024 Q4 traceability case.}
\label{fig:app-aapl-trace-audit}
\end{figure*}

\begin{figure*}[t]
\centering
\begin{tcolorbox}[colback=blue!2,colframe=blue!55,coltitle=black,colbacktitle=blue!7,title=\textbf{TSLA FY2025 Q4 Boundary Trace: Where the Forecast Is Pulled High},fonttitle=\bfseries,boxrule=0.5pt,arc=2pt,left=4pt,right=4pt,top=4pt,bottom=4pt]
\scriptsize
\setlength{\tabcolsep}{3pt}
\renewcommand{\arraystretch}{1.10}

\textbf{Panel A. Forecast path and boundary.} This boundary case starts from a high anchor; without a qualifying target-quarter guidance signal or enough downward evidence, CAME remains above the realized target.\\[2pt]
\begin{tabular}{p{0.18\linewidth}p{0.23\linewidth}p{0.51\linewidth}}
\toprule
Step & Value & Interpretation \\
\midrule
Forecast boundary & FY2025 Q3 $\rightarrow$ FY2025 Q4 & No qualifying target-quarter guidance signal under the method's taxonomy; inputs are forecast-time only. \\
Statistical Anchor & 26.33B & Selected anchor is already above the realized target before CAME corrections. \\
Evidence-memory update & Upward; conflicts unresolved & Energy demand/transition cards add upward pressure; automotive conflicts and analogs do not offset the high anchor. \\
Guarded CAME output & 27.76B CAME; 24.90B actual & The no-guidance category applies no final attenuation; guarded forecast remains above actual; actual is evaluation-only. \\
\bottomrule
\end{tabular}

\vspace{6pt}
\textbf{Panel B. Forecast-time \textit{forward-looking/guidance} evidence cards.} The upward current-evidence pressure comes mainly from Energy Generation and Storage demand/product-transition cards; automotive cards appear as conflict or lower-reliability signals but do not offset the upward path enough. The summaries are author-written and source-linked rather than verbatim. C2 concerns a longer horizon and provides only weak target-quarter support.\\[2pt]
\begin{tabular}{p{0.05\linewidth}p{0.25\linewidth}p{0.13\linewidth}p{0.17\linewidth}p{0.32\linewidth}}
\toprule
Card & Segment / factor & Signal & Role in trace & Author-written source-linked summary \\
\midrule
C1 & Energy Generation and Storage / demand & positive demand & supports upward correction (+0.1011) & Current strong demand was expected to extend into the following year. \\
C2 & Energy Generation and Storage / product transition & positive product transition & supports upward correction (+0.0674); longer-horizon, weak target-quarter support & Current positive customer feedback was reported, with shipping scheduled to begin in the following year. \\
C3 & Automotive Sales / regulatory conditions & positive regulation & conflict / lower-reliability signal (-0.0684) & Management expected removal of in-car safety drivers in Austin within a few months. \\
C4 & Automotive Sales / demand & positive demand & conflict / lower-reliability signal (-0.0518) & Management expressed confidence in expanding Tesla production. \\
\bottomrule
\end{tabular}

\vspace{6pt}
\textbf{Panel C. Forecast-time-eligible temporal memory used by the trace.} The temporal expert retrieves prior TSLA analogs with positive realized anchor-relative residuals, which pulls the temporal-memory path above the already-high anchor. Residual cues come from prior realized outcomes, not target-quarter evidence.\\[2pt]
\begin{tabular}{p{0.16\linewidth}p{0.10\linewidth}p{0.16\linewidth}p{0.50\linewidth}}
\toprule
Retrieved quarter & Weight & Residual cue & Why it is shown \\
\midrule
FY2024 Q4 & 0.4340 & +0.0206 log & Highest-weight analog; the Energy demand card matches prior deployment-growth language with match score 0.906. \\
FY2022 Q4 & 0.3512 & +0.1253 log & Positive historical residual analog that reinforces the upward temporal-memory path. \\
FY2025 Q3 & 0.2148 & +0.2732 log & Recent same-company analog with a large positive residual cue; shown despite weaker direction alignment. \\
\bottomrule
\end{tabular}

\vspace{6pt}
\textbf{Panel D. Retrospective target-quarter card-alignment check.} These rows are retrospective FY2025 Q4 target-quarter records; they record realized conditions that cut against the upward path and never enter the forecast.\\[2pt]
\begin{tabular}{p{0.18\linewidth}p{0.36\linewidth}p{0.38\linewidth}}
\toprule
Retrospective marker & Author-written target-quarter summary & Interpretation \\
\midrule
\textcolor{red!60!black}{\textbf{retrospective constraint}} & Automotive Sales / supply / negative: battery-pack availability remained the principal global constraint. & This realized automotive supply constraint cuts against the positive production/autonomy path in retrospective inspection; it is not forecast-time evidence. \\
\textcolor{red!60!black}{\textbf{monetization boundary}} & Automotive Sales / pricing / negative: FSD moved fully to a subscription-based model beginning in the target quarter. & Positive forecast-time FSD/autonomy cards did not map cleanly into near-term recognized revenue; this marks a timing/monetization boundary in retrospective inspection. \\
\bottomrule
\end{tabular}
\end{tcolorbox}
\caption{TSLA FY2025 Q4 boundary trace.}
\label{fig:app-memory-boundary-cases}
\end{figure*}

\end{document}